\documentclass[11pt]{article}

\usepackage[final]{acl}

\usepackage{times}
\usepackage{latexsym}

\usepackage[T1]{fontenc}

\usepackage[utf8]{inputenc}

\usepackage{microtype}
\usepackage{balance}
\usepackage{inconsolata}
\usepackage{subcaption}
\usepackage{caption}
\usepackage{graphicx}

\usepackage{color}
\usepackage{adjustbox}
\usepackage{listings}   
\usepackage{xcolor} 

\usepackage{amsmath,amsfonts,bm}

\def\eqref#1{equation~\ref{#1}}

\def\1{\bm{1}}

\DeclareMathAlphabet{\mathsfit}{\encodingdefault}{\sfdefault}{m}{sl}
\SetMathAlphabet{\mathsfit}{bold}{\encodingdefault}{\sfdefault}{bx}{n}

\usepackage{tcolorbox}
\tcbuselibrary{skins,breakable,fitting,xparse}
\usepackage{booktabs}
\usepackage{siunitx}
\usepackage{array}
\usepackage{multirow}
\usepackage{hyperref}
\usepackage{url}

\usepackage{enumitem}
\usepackage[normalem]{ulem}

\usepackage{booktabs}
\usepackage{tabularx}
\usepackage{ragged2e}
\newcolumntype{Y}{>{\RaggedRight\arraybackslash}X}

\usepackage{amssymb}

\usepackage{placeins}
\begin{document}
\definecolor{navyblue}{RGB}{25, 57, 138}        
\definecolor{navybluelight}{RGB}{240, 245, 255}  

\definecolor{forestgreen}{RGB}{34, 139, 34}      
\definecolor{forestgreenlight}{RGB}{240, 255, 240} 

\definecolor{burgundy}{RGB}{128, 0, 32}          
\definecolor{burgundylight}{RGB}{255, 240, 245}  

\definecolor{darkslate}{RGB}{47, 79, 79}         
\definecolor{darkslatelight}{RGB}{245, 248, 248} 

\definecolor{deeporange}{RGB}{204, 85, 0}        
\definecolor{deeporangelight}{RGB}{255, 248, 240} 

\newtcolorbox{navypromptbox}[1]{
    colback=navybluelight,
    colframe=navyblue,
    fonttitle=\bfseries\color{white},
    title=#1,
    coltitle=white,
    colbacktitle=navyblue,
    enhanced,
    attach boxed title to top left={yshift=-2mm, xshift=2mm},
    boxed title style={size=small, colback=navyblue},
    left=1mm, right=1mm, top=1mm, bottom=1mm,
    unbreakable
}

\newtcolorbox{greenpromptbox}[1]{
    colback=forestgreenlight,
    colframe=forestgreen,
    fonttitle=\bfseries\color{white},
    title=#1,
    coltitle=white,
    colbacktitle=forestgreen,
    enhanced,
    attach boxed title to top left={yshift=-2mm, xshift=2mm},
    boxed title style={size=small, colback=forestgreen},
    left=1mm, right=1mm, top=1mm, bottom=1mm,
    unbreakable
}

\newtcolorbox{burgundypromptbox}[1]{
    colback=burgundylight,
    colframe=burgundy,
    fonttitle=\bfseries\color{white},
    title=#1,
    coltitle=white,
    colbacktitle=burgundy,
    enhanced,
    attach boxed title to top left={yshift=-2mm, xshift=2mm},
    boxed title style={size=small, colback=burgundy},
    left=1mm, right=1mm, top=1mm, bottom=1mm,
    unbreakable
}

\newtcolorbox{slatepromptbox}[1]{
    colback=darkslatelight,
    colframe=darkslate,
    fonttitle=\bfseries\color{white},
    title=#1,
    coltitle=white,
    colbacktitle=darkslate,
    enhanced,
    attach boxed title to top left={yshift=-2mm, xshift=2mm},
    boxed title style={size=small, colback=darkslate},
    left=1mm, right=1mm, top=1mm, bottom=1mm,
    unbreakable
}

\newtcolorbox{orangepromptbox}[1]{
    colback=deeporangelight,
    colframe=deeporange,
    fonttitle=\bfseries\color{white},
    title=#1,
    coltitle=white,
    colbacktitle=deeporange,
    enhanced,
    attach boxed title to top left={yshift=-2mm, xshift=2mm},
    boxed title style={size=small, colback=deeporange},
    left=1mm, right=1mm, top=1mm, bottom=1mm,
    unbreakable
}

\definecolor{QuoteBorder}{RGB}{220,225,235}
\definecolor{QuoteBack}{RGB}{248,250,252}
\definecolor{QuoteGreen}{RGB}{19,87,48}
\definecolor{QuoteRed}{RGB}{128, 0, 32}

\newtcolorbox{quoteblock}[1]{%
  enhanced,
  colback=QuoteBack,
  colframe=QuoteBorder,
  boxrule=0.8pt,
  arc=6pt,
  left=8pt,right=8pt,top=3pt,bottom=3pt,
  borderline north={2.5pt}{0pt}{#1},
}

\title{Deep FinResearch Bench: Evaluating AI's Ability to Conduct Professional Financial Investment Research}


\author{Mirazul Haque $^{\dagger}$, Antony Papadimitriou $^{\dagger}$, Samuel Mensah,\\ \bf Zhiqiang Ma, Zhijin Guo, Joy Prakash Sain, Simerjot Kaur,  \\ \bf Charese Smiley \& Xiaomo Liu\\
AI Research, JPMorganChase \\
{\small \texttt{\{mirazul.haque, antony.papadimitriou, samuel.mensah, zhiqiang.ma,} } \\
  {\small \texttt{zhijin.guo, joy.sain, simerjot.kaur, charese.smiley, xiaomo.liu\}@jpmchase.com} }\\
  {\small $\dagger$Equal contribution.}
}

%

\newcommand{\fix}{\marginpar{FIX}}
\newcommand{\new}{\marginpar{NEW}}


\maketitle

\begin{abstract}
We introduce Deep FinResearch Bench, a practical and comprehensive evaluation framework for deep research (DR) agents in financial investment research. The benchmark assesses three dimensions of report quality: qualitative rigor, quantitative forecasting and valuation accuracy, and claim credibility and verifiability. Particularly, we define corresponding qualitative and quantitative evaluation metrics and implement an automated scoring procedure to enable scalable assessment. Applying the benchmark to financial reports from frontier DR agents and comparing them with reports authored by financial professionals, we find that AI-generated reports still fall short across these dimensions. These findings underscore the need for domain-specialized DR agents tailored to finance, and we hope the work establishes a foundation for standardized benchmarking of DR agents in financial research.

\end{abstract}

\section{Introduction}
Advances in generative AI, particularly large language models (LLMs), have renewed debate over whether AI can perform professional work in domains such as financial market research \cite{tomlinson2025working}. This paper evaluates AI systems in the context of professional equity research, which involves the systematic assessment of publicly traded companies to determine investment merit. Investment banks produce equity research reports to inform capital allocation decisions. High-quality research demands technical expertise in financial modeling, accounting, and valuation, as well as analytical rigor, critical thinking, clear communication, and attention to detail, all grounded in finance and capital markets theory \citep{corporate_finance_institute}.

Given the scale and importance of global equity markets, it is necessary to assess whether AI systems, particularly DR agents, can meet professional standards. Existing benchmarks are often too general \cite{du2025deepresearch} or insufficiently aligned with real-world financial research tasks \cite{sun2025finresearchbench}. A rigorous evaluation framework must combine realistic research assignments with reproducible protocols that separate retrieval, reasoning, and synthesis. To our knowledge, no widely accepted framework exists for benchmarking AI-generated equity research reports.

This paper addresses this gap by introducing a practical benchmarking framework for AI-generated equity research. Our contributions are fourfold: 
\textbf{(1)} We define three evaluation dimensions: qualitative rigor, quantitative accuracy, and claim verifiability. 
\textbf{(2)} We develop qualitative and quantitative metrics with an automated scoring mechanism. 
\textbf{(3)} We benchmark leading proprietary DR agents against professional analyst reports from two major financial institutions. 
\textbf{(4)} We conduct detailed error analysis to identify systematic failure modes and future research directions.

\begin{figure}[t]
    \centering
    \includegraphics[width=\columnwidth]{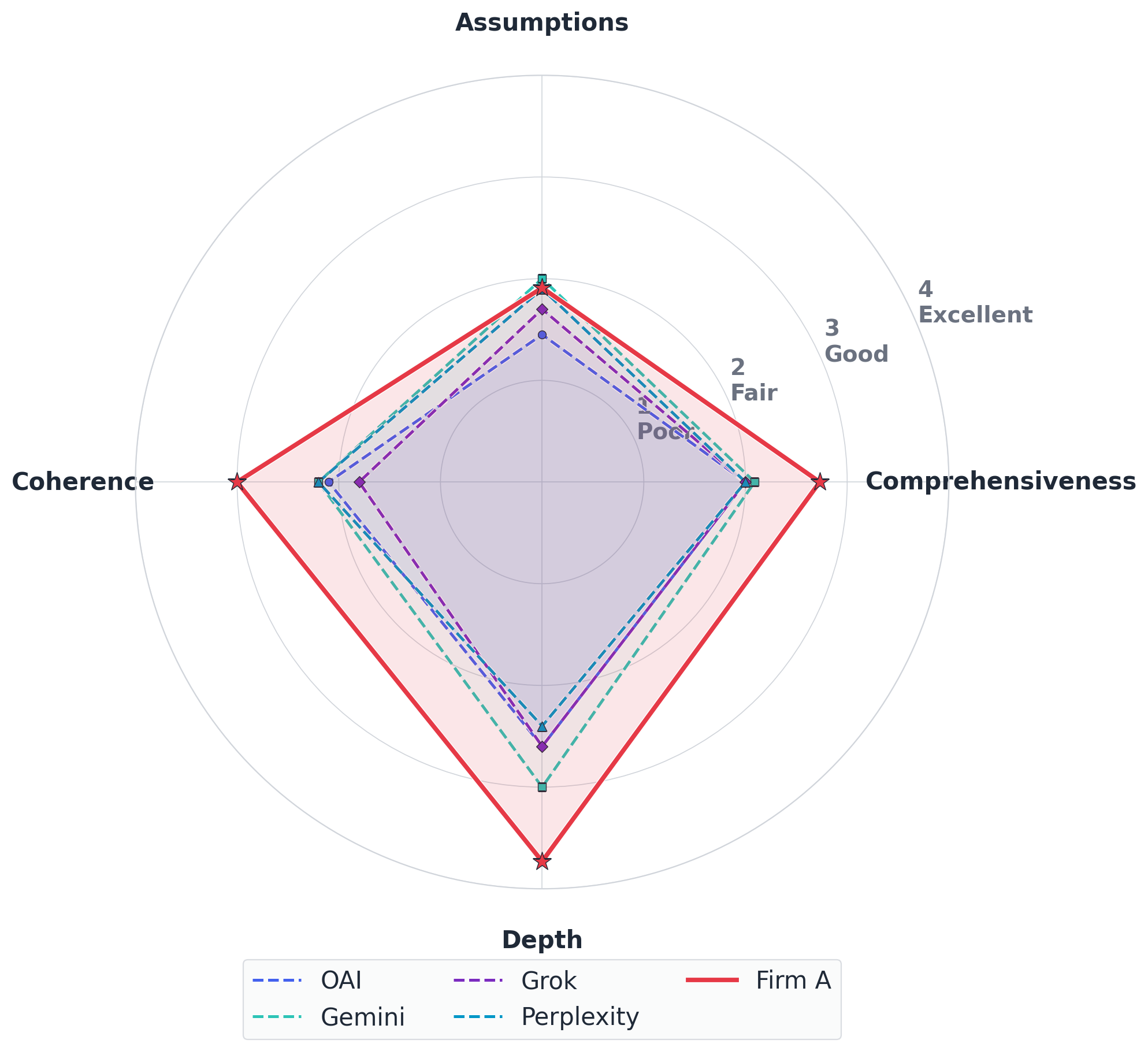}
    \caption{Qualitative performance of deep research agents and professional analysts across four evaluation dimensions (scale: 1=Poor, 2=Fair, 3=Good, 4=Excellent). Professional analysts show a clear edge over DR agents.}
    \label{fig:spider_chart}
\end{figure}

\section{Related Work}

Recently, there has been a growing trend of leveraging AI agents to automate financial analysis tasks. Trading agents have been used to conduct fundamental and technical analyses separately to develop improved stock‑trading strategies \cite{xiao2024tradingagents}. Frameworks equipped with chain‑of‑thought reasoning have been proposed to emulate human analysts for equity research \cite{zhou2024finrobot}. Additionally, proprietary LLMs are developing agent capabilities to support due‑diligence research, financial modeling, and related services.

Despite these advances, few studies rigorously evaluate whether AI agents can perform financial investment research at a professional analyst level. Related benchmarks include FinResearchBench \cite{sun2025finresearchbench} and FinDeepResearch \cite{zhu2025findeepresearch}. FinResearchBench scores report structure via an agent-as-judge logical tree, but its prompt-based reports diverge from analyst practice. FinDeepResearch writes reports section-by-section, yet emphasizes performance and competitiveness while omitting investment strategy and recommendations. Furthermore, several deep-research benchmarks~\cite{du2025deepresearch, li2026deepresearch, patel2025deepscholar} address generic tasks rather than finance-specific ones.

In contrast, our work focuses exclusively on financial investment research and provides a more effective framework for assessing AI capabilities at the professional analyst level. 

\section{Background}

\subsection{Equity Investment Research}
\label{sec:fin_research}
Equity investment research involves collecting and analyzing financial and market data to support stock-market investment and business decisions \cite{schmidt2000Investment}. Professional equity research reports typically include a company and industry overview, an investment recommendation and thesis, financial analysis and projections, and risk assessment \cite{cfainstitute2000Equity}.

High-quality research is usually produced by experienced analysts with advanced training (e.g., MBA) and certifications (e.g., CFA). Report quality depends on three core capabilities \cite{valentine2011best}: (1) \textbf{information gathering} from public sources (e.g., news) and private channels (e.g., proprietary data/reports); (2) \textbf{qualitative analysis} to identify evidence-backed drivers, scenarios, and differentiated insights that can generate alpha; and (3) \textbf{quantitative analysis} to forecast earnings and estimate valuation using reliable data and sound models, including sensitivity and scenario analysis. Further details and an example report are provided in Appendix \ref{sec:app_equity_report}.



\subsection{AI-Based Research Tools}


Producing professional-grade equity research reports (Section \ref{sec:fin_research}) motivates using generative AI, especially LLMs. Yet standard models like GPT-4 depend largely on static training data, making them inadequate for equity research that requires timely firm- and industry-specific information. Web-augmented LLMs \cite{nakano2021webgpt} help, but they are still optimized for single-step Q\&A rather than multi-stage analytical workflows.

\begin{table}[t]
\centering
\tiny
\begin{tabular}{@{}lccc}
\hline
\textbf{DR Agent} & \textbf{Model} &\textbf{Search} & \textbf{Non-search capabilities} \\
\hline

\textbf{OpenAI DR} & GPT-o3 &
Proprietary &
Code execution (Python) \\

\textbf{Gemini DR} &Gemini 2.5 Pro &
Google &
Multimodal reasoning \\

\textbf{Grok DR} & Grok 4 &
Proprietary &
Real-time Twitter data \\

\textbf{Perplexity DR} &Sonar Reasoning &
Proprietary &
Multi-document synthesis \\

\hline
\end{tabular}

\caption{
Comparison of proprietary DR agents evaluated in this study.
}
\label{tab:prop_dr_comp}
\end{table}

Deep-research (DR) agents \cite{huang2025deep} fit financial research because they combine search, planning, reasoning, and synthesis into a single pipeline (Figure \ref{fig:dr_overview}, Appendix). Financial workflows are multi-stage evidence collection, forward-looking analysis, and report writing, so single-pass LLMs remain insufficient even with search. DR systems use AI planning to decompose tasks, choose sources, run sub-analyses (e.g., EPS forecasting), and merge outputs into structured reports. Benchmarks such as DeepResearch Bench \cite{du2025deepresearch} and FinDeepResearch \cite{zhu2025findeepresearch} show clear gains over web-search-augmented LLM baselines (Table \ref{tab:llm_web_dr_comp}, Appendix).

Open-source DR frameworks (e.g., WebThinker \cite{li2025webthinker}, Tongyi DeepResearch \cite{team2025tongyi}) rely on open-source LLMs that, as of December 2025, generally trail frontier proprietary models.\footnote{See rankings at https://artificialanalysis.ai/}
 Prior results suggest proprietary DR agents outperform open-source systems \cite{zhu2025findeepresearch, han2025deer}. We therefore evaluate proprietary DR agents from OpenAI \cite{openaideepresearch}, Gemini \cite{geminideepresearch}, Perplexity \cite{perplexitydeepresearch}, and Grok \cite{grokdeepresearch} for investment research.

\section{Deep FinResearch Bench Framework}
There is no established benchmark for evaluating DR systems in professional equity research, primarily due to the lack of suitable evaluation metrics and limited access to professional research data. These constraints hinder rigorous assessment of AI systems at a professional standard. 

To address this gap, we introduce Deep FinResearch Bench, a framework for collecting professional equity research reports, generating comparable AI reports, and evaluating report quality across multiple dimensions (see Figure \ref{fig:eval_framework}). The framework enables standardized, systematic comparisons between professional analysts and AI in financial investment research.

\subsection{Professional Report Collection}
\label{sec:prof_report_gen}
The Deep FinResearch Bench pipeline begins by collecting professional equity research reports from financial institutions. As noted earlier, investment-bank reports are typically proprietary and restricted to clients, making them difficult for the public to access. To ensure reproducibility, we use a paid, publicly available Yahoo Finance subscription that provides equity research from two institutions (Firms A and B \footnote{We hide their real names due to privacy reasons. Readers can find them out through their own subscriptions.}). We then randomly sampled 25 S\&P 500 companies across three sectors, information technology, financials, and health care, and focused on fiscal year 2025, quarters Q1 and Q2. We obtained pre-earnings research reports from Firms A and B, which are typically published several weeks before quarterly results. In total, 100 professional reports were collected for evaluation.

\subsection{AI Report Generation}
\label{sec:ai_report_gen}
For a fair comparison, we generate AI reports corresponding to each professional report, using the same company, fiscal year and quarter, and closely matched publication dates. We tested two prompt-generation approaches. In the first, a simple input, specifying the target task, company, and time period, directs the DR agents to retrieve a standard equity research template and use it to produce the AI report. In the second, we first use OpenAI’s o3 model to extract detailed report rubrics for Firm A and Firm B separately; these rubrics are then embedded in the prompts as explicit guidance, enabling the DR agents to compose AI reports that closely mirror the structure of the professional reports. Our preliminary analyses show that the second approach can generate a more comprehensive analysis. Thus, we use it for a fairer comparison between professional and AI reports. Additional details on the AI report-generation process are provided in Appendix \ref{sec:ai_report_gen}. 

\subsection{Research Evaluation Methodology}

The quality evaluation on an equity research report should align with how a good professional report was written, as explained in section \ref{sec:fin_research}. After the consultations with domain experts, our proposed evaluation method concentrates on three critical aspects: qualitative analysis, quantitative analysis and verifiability \& credibility.  

\subsubsection{Metrics of Qualitative Analysis}
Good qualitative analyses should provide a thorough analysis and insightful reasoning of the investment thesis and recommendations. Different scenarios and their impacts on reasoning should be discussed. Therefore, we propose to measure the quality from the four dimensions below.  

\textbf{Comprehensiveness} checks coverage of all required sections with appropriate depth, sector-relevant KPIs, consistent citations, and minimal repetition. \textbf{Coherence} assesses logical flow, cross-references, and consistency in narrative, data, terminology, and time frames. \textbf{Assumption quality} evaluates whether assumptions are explicit, justified, specific (magnitude/horizon), aligned with the analysis, and stress-tested via ranges or scenarios. \textbf{Analytical depth} measures whether the report explains drivers and causality, addresses uncertainty and counterarguments, and yields actionable, conditional investor implications.


Detailed description of these four dimensions can be found in Appendix \ref{sec:app_qual_metrics}. The scale of each dimension ranges among poor, fair, good, excellent. They can result in more accurate and consistent grading by both AI and humans than a more granular scale, like 1-10, used in other works \cite{han2025deer, du2025deepresearch}.

\subsection{Metrics of Quantitative Analysis}
The math equations of all these metrics below can be found in Appendix \ref{sec:app_quant_metrics}.

\noindent\textbf{Financial Forecast Accuracy}. It measures the prediction of a company's future (next 1-3 years) financial performance, like revenue, EBITDA, and earnings per share (EPS), etc. We measure the error percentage between predicted and actual values using SMAPE \cite{makridakis2000m3}. 

\noindent\textbf{Stock Valuation Accuracy.} Target or fair value estimates (FVEs) expect a company to reach a future stock market price within a horizon ($h$) of 6-12 months. 
We evaluate the accuracy of FVEs produced by both AI and professional analysts using three complementary metrics.

\begin{itemize}
    \item \textit{Hit Rate} measures whether the target price was reached with a horizon $h$. 

    \item \textit{Mean Absolute Error (MAE)} captures the percentage deviation between the estimated fair value and the realized price at the end of horizon $h$.

    \item \textit{Mean Bias} measures the signed error to identify systematic optimism or conservatism at the end of $h$. Positive bias indicates optimistic targets (FVE $>$ actual); negative bias indicates conservative targets.
\end{itemize}

\noindent\textbf{Investment Recommendation Accuracy.} Based on the fair value estimates and proposed investment thesis, a report also makes recommendations of BUY, SELL, and HOLD options to customers. Its accuracy is calculated by the following two metrics. 

\begin{itemize}
    \item \textit{Directional Accuracy (DA)} evaluates whether realized price movement after the horizon $h$ aligns with the investment recommendation. Considering only actionable signals $s \in \mathcal{S} = \{\text{BUY}, \text{SELL}\}$ (HOLD signals are excluded as they imply no directional prediction).

    \item \textit{Signed Recommendation Loss (SRL)} is a magnitude-weighted loss function that penalizes incorrect directional signals based on realized excess returns. Lower SRL indicates superior recommendation quality and economic alignment. 
\end{itemize}


\subsection{Verifiability \& Credibility Metrics}
For verifiability evaluation, we use a fine-grained, claim-level protocol. Our objective is to measure three metrics: factuality rate ($F(R)$), hallucination rate ($H(R)$), and non-verification rate ($NV(R)$).Given a report $R$ produced by DR agent, we define factuality as the proportion of claims that can be externally verified. $NV(R)$ shows the proportion of claims that can't be externally verified, while $H(R)$ represents the claims where the claim is not present in the given citations.

As shown in Figure \ref{fig:eval_framework}, the pipeline has three stages: (i) an LLM extracts claims and citations from the report, (ii) an LLM with web access verifies each claim independently, and (iii) we aggregate the verdicts to obtain $F(R)$ as the report’s overall factuality.

We further categorize claims as numeric or descriptive. A numeric claim is a declarative statement verifiable (or falsifiable) by checking one or more explicit measurable values (e.g., numbers, currencies, percentages, per-share amounts, dates, or counts) tied to a specific entity and time period. A descriptive claim is a declarative statement whose truth depends primarily on non-numeric attributes, categorical conditions, or relations (e.g., status, compliance, policy changes, risk statements, records/superlatives, dependencies, or legal/organizational actions) without requiring a numeric value to evaluate correctness.

\subsection{LLM Judge \& Evaluation}
Quantitative and verifiability metrics are objective and can be benchmarked against ground truth. By contrast, qualitative metrics are largely subjective, and the most reliable ground truth comes from expert manual grading. In our preliminary analysis, grading each report required approximately tens of hours per annotator. To scale the evaluation, we developed separate prompts for each of four qualitative dimensions and adopted an “LLM-as-a-judge” approach \cite{zheng2023judging}. The full judge prompts, for comprehensiveness, coherence, assumption quality, and analytical depth, are provided in Appendix \ref{sec:app_judge_prompt}, and key prompt-engineering considerations for improving automated grading are discussed in Appendix \ref{sec:app_llm_judge}.

\begin{table}[htbp]
\centering
\scriptsize

\begin{tabular}{@{}lccc}
\hline
 \textbf{Report} & \textbf{GPT-5} & \textbf{Gemini 2.5 Pro} & \textbf{Sonnet 4} \\ 
  
\hline

 OpenAI  & \textbf{75.0} & \underline{45.0} & 30.0 \\
 Gemini  & \textbf{85.0} & \underline{45.0} & \underline{45.0} \\
 Grok    & \textbf{85.0} & \underline{65.0} & 50.0 \\
 Perp.   & \textbf{75.0} & \underline{60.0} & 0.0 \\
 Overall & \textbf{80.0} & \underline{53.8} & 31.2 \\
\hline
\end{tabular}
\caption{Pairwise agreement rate between LLM judges and human annotations on qualitative evaluation of DR reports.}
\label{tab:agreement_metrics}
\end{table}

For LLM-based judging, we tested frontier LLMs, OpenAI GPT-5, Gemini 2.5 Pro, and Claude Sonnet 4, as judges \footnote{We did not include Grok due to poor evaluation quality. Perplexity relies on open-source LLMs that are not yet frontier.}. To ensure the quality of LLM judges, we first assessed human-LLM alignment in a small-scale evaluation. We randomly selected eight companies and collected four research reports for each. Eight domain experts annotated the qualitative metrics using the same guidance provided to the LLM judges, ensuring that each LLM‑judged report had at least two human annotators. We then evaluated judge quality using a pairwise agreement rate; the proportion of report pairs in which human and LLM grading matched \cite{han2025deer}. As shown in Table \ref{tab:agreement_metrics}, GPT‑5 performed best and aligned closely with human experts. Accordingly, we selected GPT‑5 as the primary LLM judge for subsequent experiments (we also calculated Pearson and Spearman correlations and reached the same conclusion. Please see Appendix \ref{sec:app_llm_judge}).

\section{Experiments and Results}
Due to limited space, we present evaluation results for Firm A reports in this section. The results on Firm B reports are shown in Appendix \ref{sec:app_firmb_eval}. Also, the comparison between LLMs with web search and DR agent are introduced in Appendix \ref{sec:app_llm_web_eval}. 

\subsection{Qualitative Evaluation}

Table \ref{tab:model_performance} shows a clear separation between professional analysts and DR agents, alongside meaningful variation across agent designs. Firm~A achieves the highest scores in comprehensiveness, coherence, depth, and overall quality, with depth (3.73) and coherence (3.00) approaching the ``good'' range on the four-point scale.

Among agents, Gemini is the strongest overall performer, with the highest aggregate score (2.31) and second-place rankings behind Firm~A across the major qualitative dimensions. It also attains the highest assumptions score (2.00) across all entries, including professional analysts, indicating comparatively stronger methodological transparency.

Perplexity ranks second among agents and performs competitively on coherence and assumptions but trails Gemini in depth and overall quality. OpenAI demonstrates moderate coherence and depth yet records the lowest assumptions score, reflecting weaker articulation of modeling premises. Grok performs less strongly overall and does not lead any qualitative dimension.

In summary, DR agents consistently produce reports of ``fair'' quality and, in specific areas such as methodological assumptions, can match or exceed human analysts. However, professional analysts retain a clear advantage in higher-order qualities, particularly narrative coherence, analytical depth, and overall integration.

\begin{table}[t]
\centering
\scriptsize
\begin{tabular}{@{}lccccc}
\hline
\textbf{Report} & \textbf{Comp.} & \textbf{Assum.} & \textbf{Cohe.} & \textbf{Dep.} & \textbf{Overall} \\
\hline
\multicolumn{6}{c}{Deep Research Agents} \\
\hline
OAI        
& 2.00 
& 1.45 
& 2.10 
& 2.60 
& 2.02 \\
Gemini     
& \uline{2.09} 
& \textbf{2.00} 
& \uline{2.20} 
& \uline{3.00} 
& \uline{2.31} \\

Grok       
& 2.00 
& 1.70 
& 1.80 
& 2.60 
& 2.02 \\

Perplexity 
& 2.00 
& 1.90 
& \uline{2.20} 
& 2.40 
& 2.12 \\

\hline
\multicolumn{6}{c}{Professional Analysts} \\
\hline
Firm A 
& \textbf{2.73} 
& \uline{1.91} 
& \textbf{3.00} 
& \textbf{3.73} 
& \textbf{2.84} \\
\hline
\end{tabular}

\caption{Model performance statistics showing mean scores across metrics (scale: 1=Poor, 2=Fair, 3=Good, 4=Excellent). Bold indicates best and underlined indicates second-best within each metric.}
\label{tab:model_performance}
\end{table}


\subsection{Quantitative Evaluation}

Table \ref{tab:forecast_accuracy_smape_all} reveals a clear performance hierarchy in financial forecasting. Firm~A achieves the lowest aggregate SMAPE and leads on revenue, EBITDA, operating income, and net income. Among DR agents, Grok delivers the strongest overall accuracy, ranking second on most income-statement measures, while Gemini excels on free cash flow and EPS. Although DR agents match analyst-level accuracy on selected metrics, professional analysts consistently outperform on core income-statement forecasts.

A different pattern emerges for stock-level outcomes (Tables \ref{tab:target_price_metrics} and \ref{tab:stock_rec_metrics}). Directional leadership varies by horizon, Gemini leads at 3 months and OpenAI at 6 months, while Firm~A demonstrates superior valuation calibration, achieving the lowest short-horizon SRL and mean bias at both horizons. DR agents substantially narrow the gap with human analysts on recommendation and valuation tasks, particularly at medium horizons, but professional analysts remain more consistent across financial statement forecasting.


\begin{table}[t]
\centering
\scriptsize

\begin{tabular}{@{}lccccc}
\hline
\textbf{Metric}
& \textbf{OAI}
& \textbf{Gemini}
& \textbf{Grok}
& \textbf{Perp.}
& \textbf{Firm A} \\
\hline

Revenue
& 24.73
& 22.63
& \uline{18.65}
& 23.92
& \textbf{10.64} \\

EBITDA
& 12.13
& 22.52
& \uline{10.82}
& 38.60
& \textbf{6.58} \\

Operating Income
& \uline{16.78}
& 29.18
& 19.27
& 29.90
& \textbf{5.32} \\

Net Income
& 16.62
& 21.58
& \uline{14.32}
& 18.76
& \textbf{7.43} \\

Free Cash Flow
& 18.54
& \textbf{9.51}
& \uline{12.58}
& 15.64
& 51.93 \\

EPS
& 39.06
& \textbf{27.23}
& \textbf{27.23}
& 33.60
& \uline{27.64} \\

\hline
Overall
& 21.52
& 23.05
& \uline{17.49}
& 27.56
& \textbf{17.14} \\

\hline
\end{tabular}

\caption{
Forecast accuracy measured by SMAPE (\%).
Lower values indicate better performance.
Bold indicates best and underline indicates second-best result per metric
}
\label{tab:forecast_accuracy_smape_all}
\end{table}

\begin{table}[t]
\centering
\scriptsize
\begin{tabular}{@{}lcccc@{}}
\hline
\textbf{Report} & \textbf{Horizon} & \textbf{Hit Rate (\%)} & \textbf{MAE (\%)} & \textbf{Mean Bias (\%)} \\
\hline
\multirow{2}{*}{OAI}
& 3m & \textbf{33.3} & 26.4 & \uline{+10.0} \\
& 6m & 46.2 & \uline{27.4} & \uline{+4.3} \\
\hline
\multirow{2}{*}{Gemini}
& 3m & \uline{31.8} & 28.2 & +17.0 \\
& 6m & \textbf{57.1} & 28.0 & +8.4 \\
\hline
\multirow{2}{*}{Perplexity}
& 3m & 29.3 & \textbf{22.3} & +19.7 \\
& 6m & \uline{56.4} & \textbf{19.1} & +11.0 \\
\hline
\multirow{2}{*}{Grok}
& 3m & 20.0 & 50.3 & +33.6 \\
& 6m & 44.1 & 49.3 & +25.1 \\
\hline
\multirow{2}{*}{Firm A}
& 3m & 25.8 & \uline{24.4} & \textbf{+8.0} \\
& 6m & 40.0 & 30.2 & \textbf{+3.6} \\
\hline
\end{tabular}

\caption{Target price metrics by horizon. Hit rate: higher is better. MAE: lower is better. Mean bias ranking uses absolute magnitude (closer to 0 is better) while retaining sign. Bold indicates best and underlined indicates second-best within each horizon and metric.}
\label{tab:target_price_metrics}
\end{table}

\begin{table}[t]
\scriptsize
\centering
\begin{tabular}{@{}lccc}
\hline
\textbf{Report} & \textbf{Horizon} & \textbf{Dir. Acc. (\%)} & \textbf{SRL} \\
\hline
\multirow{2}{*}{OAI}
& 3m & 63.3 & 0.045 \\
& 6m & \textbf{76.9} & \textbf{0.052} \\
\hline
\multirow{2}{*}{Gemini}
& 3m & \textbf{72.7} & \uline{0.037} \\
& 6m & 61.9 & \uline{0.060} \\
\hline
\multirow{2}{*}{Perplexity}
& 3m & \uline{68.3} & 0.051 \\
& 6m & \uline{71.8} & 0.061 \\
\hline
\multirow{2}{*}{Grok}
& 3m & 51.4 & 0.071 \\
& 6m & 61.8 & 0.088 \\
\hline
\multirow{2}{*}{Firm A}
& 3m & 64.5 & \textbf{0.036} \\
& 6m & 53.3 & 0.070 \\
\hline
\end{tabular}
\caption{Stock recommendation metrics by horizon. Directional accuracy: higher is better. SRL: lower is better. Bold indicates best and underlined indicates second-best within each horizon and metric.}
\label{tab:stock_rec_metrics}
\end{table}



\subsection{Verifiability \& Credibility Evaluation}
\noindent\textbf{Verifiability.} To assess verifiability, we use a two-stage, claim-centric pipeline. First, we extract atomic claims from system outputs using GPT-5. Second, we verify each claim with LLMs via a two-step procedure: (i) for each claim-linked URL, we extract page content and use GPT-4.1 to verify the claim against that content; (ii) if extraction fails, we use GPT-4o with web search to retrieve the cited sources and compare them to the claim.

We report (i) mean number of generated claims, split into numeric and descriptive, and (ii) claim-level factuality rates. The factuality rate $F(R)$ is percentage of claims supported by their cited sources, reported overall and by claim type. We also report hallucination rate $H(R)$ and non-verification rate $NV(R)$, where $NV(R)$ captures cases with generic, dynamic, or inaccessible links. Table~\ref{tab:hallucination_summary} summarize results across four DR agents (OpenAI, Gemini, Grok, Perplexity) for Firm A.

Across systems, the mean claim counts are similar, with Gemini highest on average, and numeric claims exceeding descriptive claims for all systems. OpenAI and Perplexity achieve the highest overall $F(R)$, while Gemini shows lower factuality despite strong qualitative outputs. $NV(R)$ is higher for Gemini and Grok, indicating more generic or dynamic citations. For Firm B, evaluation can be found in Table~\ref{tab:hallucination_summary_arg} and in Section \ref{sec:app_supp_eval}(Appendix).

In addition to evaluating the verifiability of DR-generated reports, we also evaluated the verifiability of human reports. The discussion can be found in Section \ref{sec:human_report}.

\noindent\textbf{Credibility.} We also measure the credibility of the web-sources cited by DR agents. Detailed evaluation can be found in and in Section \ref{sec:app_supp_eval} (in Appendix).

\begin{table}[htbp]
\centering
\scriptsize
\setlength{\tabcolsep}{6pt}

\begin{tabular}{@{}lcccc}
\hline 
 \textbf{Metrics} & \textbf{OpenAI} & \textbf{Gemini} & \textbf{Grok} & \textbf{Perp.} \\
 \hline
 \# claims        &  23.4   & 27.4   & 19.1   & 24.7 \\
 \# num. claims   &  18.7   & 22.3   & 15.7   & 20.6 \\
 \# desc. claims  &  4.8    & 5.4    & 3.6    & 4.3  \\
 \hline
overall $F(R)$    &  \textbf{86.0\%} & 69.6\% & 53.2\% & \uline{75.6\%} \\
num. $F(R)$       &  \textbf{84.9\%} & 68.8\% & 50.2\% & \uline{73.7\%} \\
desc. $F(R)$      &  \textbf{89.4\%} & 75.4\% & 61.9\% & \uline{88.7\%} \\
overall $H(R)$    &  \textbf{11.2\%} & \uline{17.3\%} & 34.1\% & 18.5\% \\
overall $NV(R)$   &  \textbf{2.9\%}  & 13.1\% & 12.7\% & \uline{5.9\%} \\
\hline
\end{tabular}

\caption{Verifiability analysis (mean values) of DR agent-generated reports where the generated reports follow the structure of Firm A. Bold indicates best and underlined indicates second-best within each metric.}
\label{tab:hallucination_summary}
\end{table}

\subsection{Summary and Analysis}
Overall, the evaluation results on both Firm A and B reports showed a consistent pattern that professionals can still perform better equity investment research than the best AI research tool - DR agents. We conducted some further investigations and case studies on where DR agents still lag behind in Appendix \ref{sec:prof_vs_dr_analyses}. We believe these analyses can provide useful insights of future enhancements to DR agents in financial research. 



\section{Conclusion}
This work introduces Deep FinResearch Bench framework to rigorously and reproducibly assess whether DR agents can produce professional‑grade equity research, pairing professional reports with AI outputs and grading qualitative, quantitative, and factuality dimensions. Across four leading proprietary agents, we find that AI reports are well‑structured and numerically explicit but consistently underperform professional analysts: qualitatively, they lack sector‑specific KPIs, robust assumption justification, and scenario/sensitivity analysis; quantitatively, their forecasting (e.g., revenue, EPS) trails professional and consensus baselines. In addition, agent generated claims also suffer from significant hallucinations. We aim for FinResearch Bench to serve as a practical blueprint toward trustworthy, decision‑relevant AI research systems in finance.

\noindent\textbf{Disclaimer.} This paper was prepared for informational purposes by the Artificial Intelligence Research group of JPMorgan Chase \& Co. and its affiliates (“JP Morgan’’) and is not a product of the Research Department of JP Morgan. JP Morgan makes no representation and warranty whatsoever and disclaims all liability, for the completeness, accuracy, or reliability of the information contained herein. This document is not intended as investment research or investment advice, or a recommendation, offer or solicitation for the purchase or sale of any security, financial instrument, financial product, or service, or to be used in any way for evaluating the merits of participating in any transaction, and shall not constitute a solicitation under any jurisdiction or to any person, if such solicitation under such jurisdiction or to such person would be unlawful. © 2026 JPMorganChase \& Co. All rights reserved.

\bibliography{iclr2026_conference}

\clearpage
\appendix
\section{Appendix}
\subsection{Limitations}
Deep research agents are advancing at a fast pace. This study mainly explored the commercial deep research agents, which is more likely adopted in business uses of big corporations, for financial investment research. We didn't fully explore the options of open source deep research agents like DeerFlow \cite{bytebance2025deerflow} and OpenManus \cite{liang2025openmanus} etc. These projects disclose more details of their implementations of deep research agents. There are more opportunities to observe the insights of producing AI research reports from them even if their performances may not be frontier comparing to the commercial offerings.    

Another limitation is the scope of financial research tasks investigated in this paper. Equity investment research is just one type of financial investment research tasks. The capabilities of AI systems like deep research agents on other tasks like credit, commodity and macro-economy research remain unexplored, largely because equity research is much more popular and consumed by both institutional and retail customers. But other asset classes also play critical roles in the financial market. They may require slightly different skills of qualitative and quantitative analyses which may require new architecture of deep research agents. They are our future research directions.

\subsection{Professional Equity Research Report}
\label{sec:app_equity_report}

Equity investment research is the process of gathering and analyzing financial and market data to make informed investment and business decisions on equity asset (other asset types can be credit, commodity and foreign  exchange etc.) \cite{schmidt2000Investment}. It is a critical skill set for financial professionals to master in the investment banking business. Many investment banks have a global research department with dedicated teams to conduct insightful equity investment research, generate research reports and provide guidance to investment opportunities. Most of professional equity research reports should include the following components \cite{cfainstitute2000Equity}:
\begin{itemize}
    \item \textbf{Company \& Industry Overview} - a report usually starts with a detailed description of the company, its business model and competitive position in its industry.
    \item \textbf{Investment Recommendation} - it highlights financial analysts' projection of the company stock's future value typically in next 12 months and their recommendations to investors on buying, holding, or selling stock shares of the company.
    \item \textbf{Investment Thesis} - a summary of the reasoning and argument on why analysts have make the proposed investment strategy and recommendation. The entire report should be consistent with the thesis.  
    \item \textbf{Financial Analysis \& Projection} - evaluate the financial performance of a company using their financial statements and key valuation metrics and forecast its future earnings and then valuation of its stock.  
    \item \textbf{Investment Risks} - the potential company, industry and macroeconomic challenges that may poise risks to the investment thesis.
\end{itemize}

\begin{figure*}[ht]
    \centering
    \includegraphics[width=0.95\textwidth]{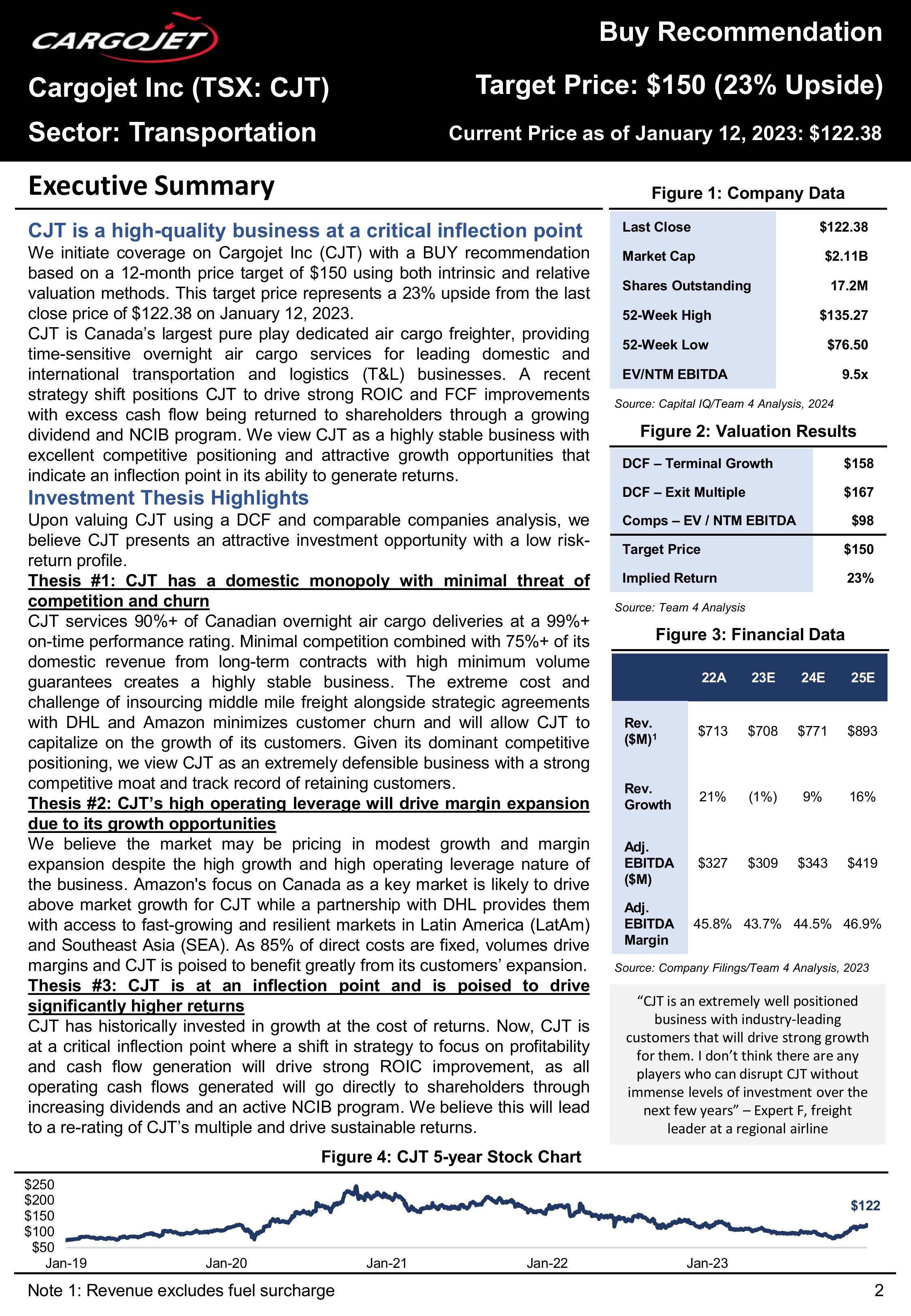}
    \caption{First page of a  mock up equity research report from the global final champion of CFA Institute Research Challenge in 2024. This report is quite similar to the real research reports written by professional analysts.}
    \label{fig:human_report_example1}
\end{figure*}

Most of equity research reports produced by investment banks are only visible to their customers. A report can be several to tens of pages long depending on the amount of company and industry developments needed for the analysis. Figure \ref{fig:human_report_example1} shows first page of a public example research report \footnote{from a CFA Institute research challenge champion, https://tinyurl.com/2akhrp3t}. 

In financial services industry, a successful equity research analyst may require advanced education (e.g. MBA), professional certification (e.g. CFA) and multi-years work experience. The quality of an equity research report depends on three types of analytical capabilities as follows \cite{valentine2011best}: 

\begin{itemize}
    \item \textbf{Information Gathering} to discover and filter key insights about the company and sector from public sources (e.g. news) and private channels (e.g. proprietary data \& reports).

    \item \textbf{Qualitative Analysis} to identify and monitor critical factors and scenarios of impacting financial performance and valuation of a company. The good analysis should unveil insights distinctive from the market consensus. Because generating alpha from consensus is a key approach of generating return from equity investment.     

    \item \textbf{Quantitative Analysis} to create financial forecast of a company future earnings and stock valuations based on its historical financials, competitiveness, industry trend and macroeconomic environment. The high quality analysis should use reliable data, sound financial modeling technique and include sensitive and scenarios analyses to cover different outcomes due the changes in financial assumptions and market environments. 
\end{itemize}
 
We designed the FinResearch evaluation metrics to measure the quality of a report based on the required key components and analytical capabilities.

\begin{figure*}[ht]
    \centering
    \includegraphics[width=0.95\textwidth, angle=90]{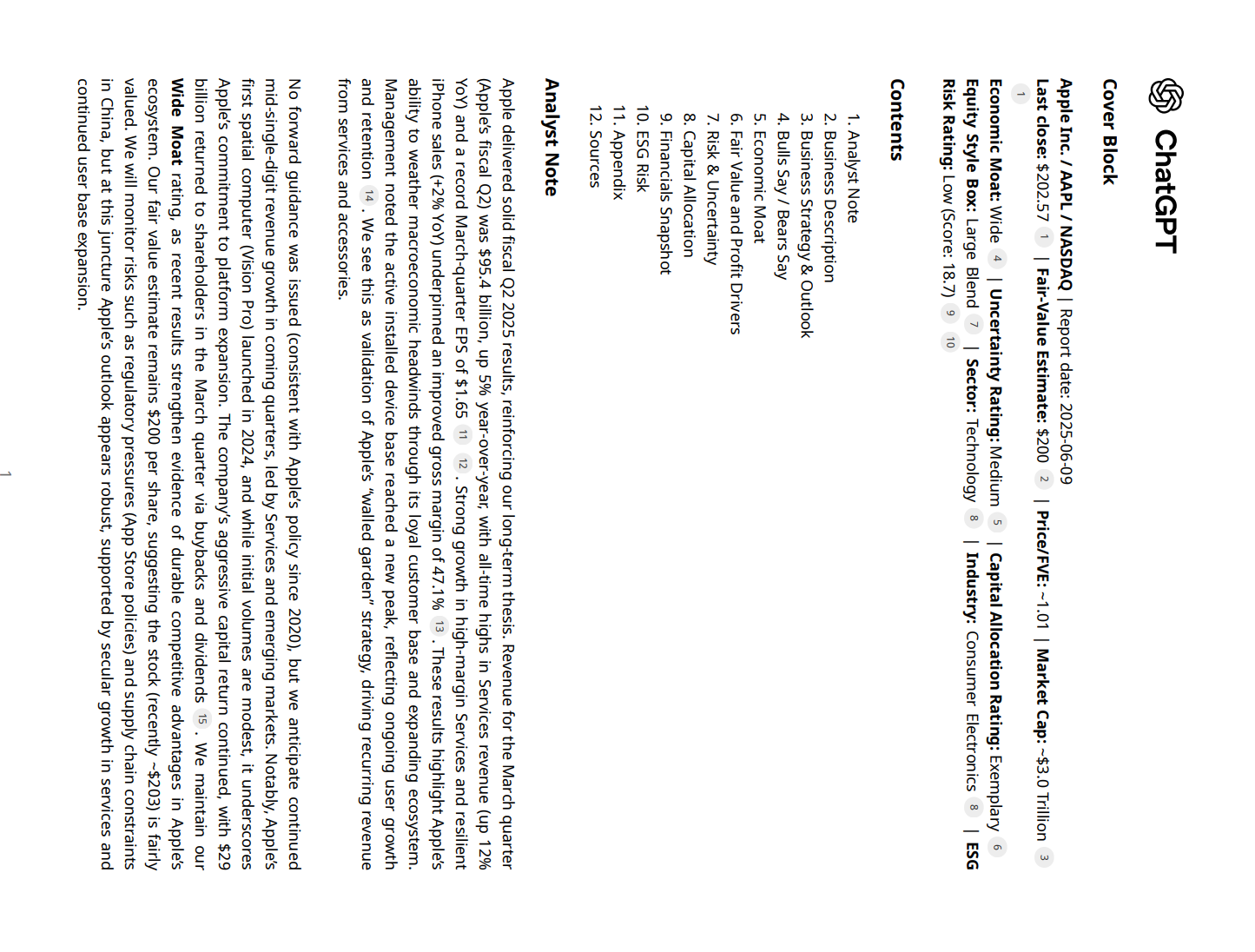}
    \caption{First page of an example of equity research report generated by OpenAI deep research agent using the detailed prompt with the rubrics extracted Firm A.}
    \label{fig:ai_report_example1}
\end{figure*}

Figure \ref{fig:human_report_example1} shows the cover page of an example of professional equity research report. It was written by students of University of Waterloo, Canada which is the global final and Americas regional champions of CFA Institute research challenge \footnote{https://www.cfainstitute.org/insights/events/research-challenge/past-champions}. This report is close to the structure and content of research reports published by renowned financial institutions. The full report can be found here \footnote{https://www.cfainstitute.org/sites/default/files/-/media/documents/support/research-challenge/challenge/rc-2024-winning-written-report-university-of-waterloo.pdf}.

Figure \ref{fig:ai_report_example1} shows the cover page of an equity research report generated by OpenAI Deep Research feature with ChatGPT application. This report was generated on 06/09/2025 for the outlook of Apple Inc in Q4 2025.

\subsection{AI-Based Research Tools}

Some evidences have already emerged to exhibit the advantages of DR agents over single-pass LLMs (even with web search) on research tasks. OpenAI revealed that its DR agent can score 26.6\% accuracy on Humanity's Last Exam benchmark, which consists of expert-level problem solving tasks to test deep reasoning and broad knowledge, while the best LLMs at the time only can only achieve around 10\% at a single pass \cite{openaideepresearch} \footnote{Results were evaluated on Feb. 2025}.There are also deep research benchmarks including Deep Research Bench \cite{du2025deepresearch} and FinDeepResearch \cite{zhu2025findeepresearch} indicating salient performance boost over the method of using base LLMs with web search only. Table \ref{tab:llm_web_dr_comp} below summarizes the performance comparisons between LLMs with web search and deep research agents on DeepResearch Bench and FinDeepResearch benchmarks.

\begin{figure*}[t]
    \centering
    \includegraphics[width=0.9\linewidth]{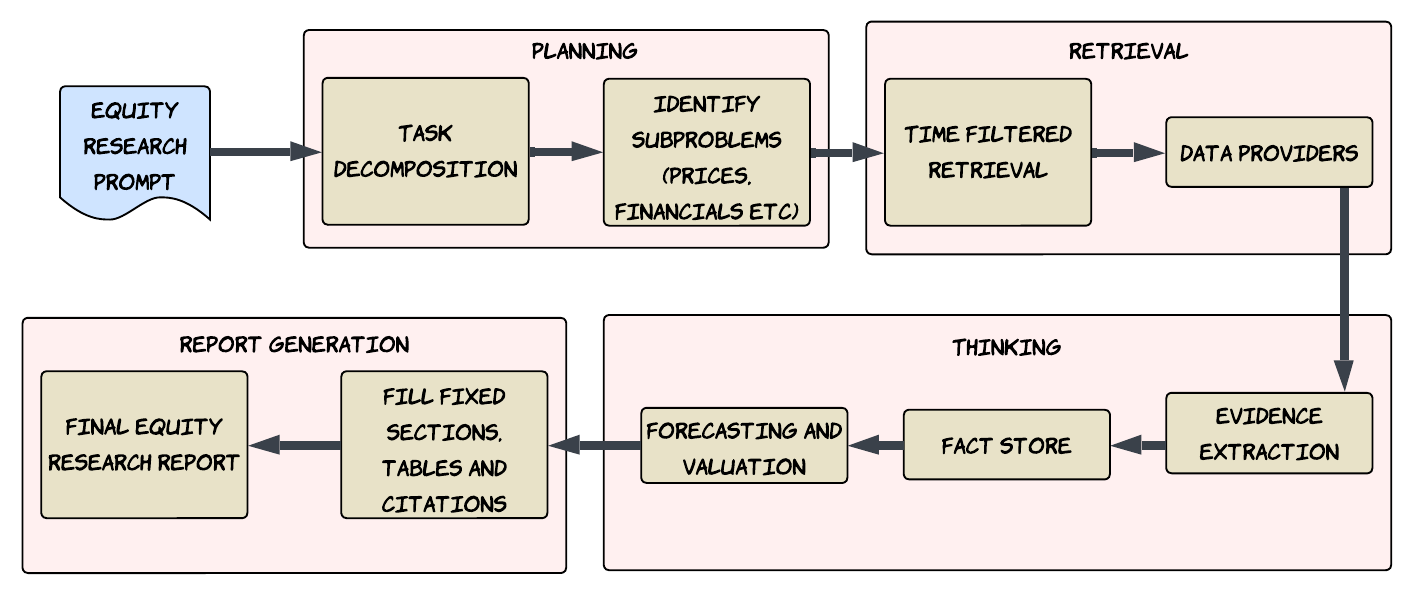}
    \caption{Overview of deep research framework.}
    \label{fig:dr_overview}
\end{figure*}

\begin{table}[htbp]
\centering
\tiny
\setlength{\tabcolsep}{6pt}
\renewcommand{\arraystretch}{1.05}
\begin{adjustbox}{width=\columnwidth}
\begin{tabular}{ccccccc}
\toprule
\textbf{Model} &  \multicolumn{2}{c}{\textbf{DeepResearch Bench}} & \multicolumn{3}{c}{\textbf{FinDeepResearch}}  \\
\cmidrule{2-3} \cmidrule{4-6}
             & RACE  & FACT   & Precision & Rigor & Acc. \\
\midrule
\multicolumn{6}{c}{LLM with Web Search} \\ 
\midrule
GPT-4.0 / 5.0    & 35.1 & \underline{88.4} & 33.0  & \underline{99.9} & 37.4 \\
Gemini 2.5 Pro   & 35.1 & 81.8 & 24.4  & 99.3 & 22.9 \\
Sonar Reasoning  & 40.2 & 48.7 & -- & -- & -- \\
Grok 4           & --   & --   & 22.1 & \underline{99.9} & 23.7 \\
\midrule
\multicolumn{6}{c}{Deep Research Agent}     \\
\midrule
OpenAI DR       & \underline{47.0} & 78.0 & \textbf{37.9} & 99.4 & \textbf{42.5} \\ 
Gemini DR       & \textbf{48.9} & 81.4 & 36.2 & -- & \underline{37.6} \\
Grok DR         & 40.2 & 83.6 & \underline{37.3} & -- & 34.5 \\
Perplexity DR   & 42.3 & \textbf{90.2} & 24.7 & \textbf{100} & 21 \\
\bottomrule
\end{tabular}
\end{adjustbox}

\caption{Performance results of AI-based research tools on deep research benchmarks.}
\label{tab:llm_web_dr_comp}
\end{table}

\subsection{Deep FinResearch Bench Details}

\begin{figure*}[t]
    \centering
    \includegraphics[width=\linewidth]{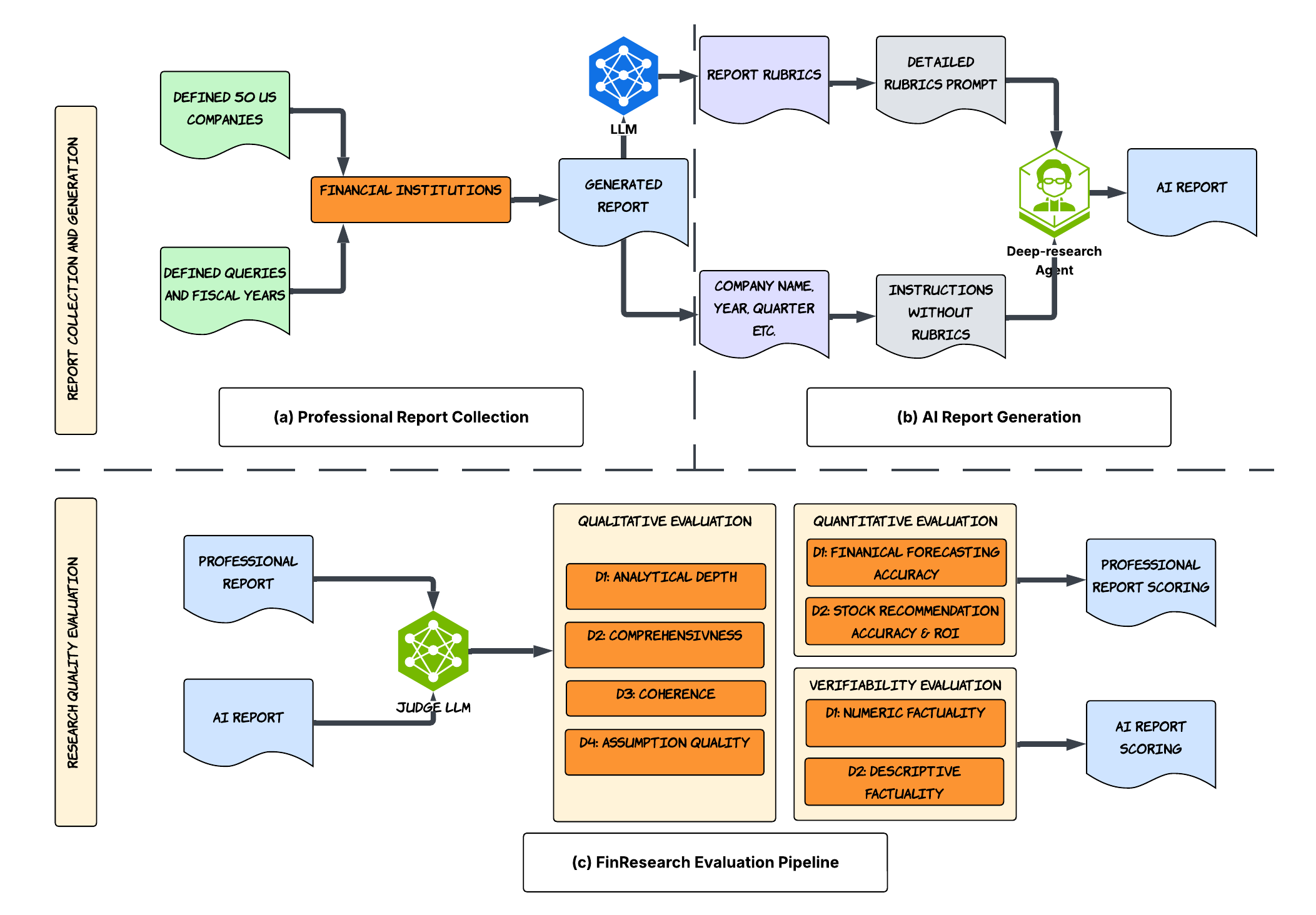}
    \caption{Overview of evaluation framework}
    \label{fig:eval_framework}
\end{figure*}

\subsubsection{AI Report Generation}
\label{sec:app_ai_report_gen}

AI-generated reports typically include sections such as analyst note, business description, investment recommendations, investment thesis, and risk drivers, mirroring professional equity research reports; see the example in Appendix \ref{fig:ai_report_example1}. Based on OpenAI Deep Research (DR) logs, the agent decomposes the task into sub-tasks: quantitative (historical financial statements, forward-looking forecasts, valuation) and qualitative (competitive moat, key risks, capital allocation), each mapped to explicit data requirements (e.g., revenue, EPS, ratings). 

For convenient experimentation, we retrospectively generated all the AI reports after the corresponding professional report was published, and sometimes the official earnings were released as well. We then imposed a knowledge cut-off constraint in the prompts to ensure the DR agents only use and cite information on and before the publication date of the comparing professional reports. The full prompts can be seen in Appendix \ref{sec:app_judge_prompt}. The system then issues targeted web searches across news, filings, and other sources, adhering to the knowledge cutoff constraint. We analyzed all 100 AI reports and verified nearly 100\% citation compliance with this constraint. The AI research process proceeds through an iterative loop, plan → search → read → extract → synthesize → verify, as illustrated in the Deep Research agent framework (Figure \ref{fig:dr_overview})

\subsubsection{Metrics of Qualitative Analysis}
\label{sec:app_qual_metrics}

\textbf{Comprehensiveness} evaluates whether a report meets expectations for content coverage. It examines whether all required areas are addressed with appropriate depth and sector-specific detail. A comprehensive report includes the expected sections: executive summary, business description, outlook and strategy, valuation, and risk drivers, and provides proportional depth aligned to each section’s importance within the overall analysis. It should also incorporate sector-appropriate key performance indicators, using industry-specific metrics such as net interest income or net interest margin for banks, annual recurring revenue or net revenue retention for SaaS companies, same-store sales for retail, and production volumes for the energy sector. Evidence should be integrated effectively, with consistent source citation to support claims throughout. Finally, redundancy should be minimized by avoiding repetition across sections unless it adds new depth or perspective.

\textbf{Coherence} is the extent to which a report presents a connected, unified argument, ensuring the narrative reads as an integrated whole rather than a set of disconnected sections. A coherent report demonstrates clarity and flow, using precise language and smooth transitions within and across sections, with ideas progressing logically throughout. Structural coherence is achieved when sections reference and reinforce one another, through explicit cross-references that link assumptions, valuations, and risks, rather than presenting information as isolated lists. Internal consistency is essential: there should be no contradictions among the narrative, tables, assumptions, dates, or cited facts. Consistent use of defined terminology and jargon further supports coherence, as does temporal accuracy, where time references (e.g., “recent,” “now”) align with the actual dates cited in the report.

\textbf{Assumption} quality is the rigor and transparency of the premises underpinning a report’s forecasts, valuations, and conclusions. High-quality assumptions are stated explicitly and placed near the forecasts they inform, rather than buried or implied. Each assumption should be thoroughly justified using multi-year historical data, peer or industry benchmarks, or clearly cited sources. Specificity is essential: define magnitudes, units, time horizons, and causal drivers, rather than rely on vague generalities. Consistency must be maintained so assumptions align with the report’s tables, figures, and overall conclusions. Robust assumption quality is demonstrated through stress-testing, presenting material drivers with quantified ranges or scenario analyses, rather than relying solely on single-point estimates.

\textbf{Analytical depth} is the extent to which a report moves beyond description to deliver substantive insight. A report with strong analytical depth not only explains what happened but also clarifies why, offering causal explanations for observed outcomes. Inference quality is highest when forecasts or hypotheses are explicitly linked to identifiable drivers. Effective use of data includes quantified assumptions and benchmarks, alongside a clear distinction between correlation and causation. Analytical depth also involves addressing counterpoints and uncertainty, exploring alternative scenarios, sensitivities, limitations, and competing explanations. Finally, reports that demonstrate analytical depth deliver actionable implications, articulating clear, conditional takeaways directly relevant to investor decision-making.

\subsubsection{Metrics of Quantitative Analysis}
\label{sec:app_quant_metrics}

\textbf{Financial Forecast Accuracy}. The symmetric mean absolute percentage error (SMAPE) is defined as follows. Given predicted the value $F_i$ and
actual value $A_i$ of a financial item (e.g. revenue, EBITDA, EPS) for company $i$, SMAPE is the mean of the symmetric absolution percentage errors of a set of companies as
\begin{equation}
\text{SMAPE} = \frac{100\%}{n} \sum_{i=1}^{n} \frac{|A_i - F_i|}{\frac{|A_i| + |F_i|}{2}}
\end{equation}
where $n$ is the number of companies in the evaluation.

\noindent\textbf{Stock Valuation Accuracy}.
Each fair value estimate (FVE) is mapped to an investment signal based on its ratio to the
contemporaneous market price $P_t$:
\begin{equation}
r_t = \text{FVE} / P_t
\end{equation}
Signals are assigned using fixed thresholds:
\begin{equation}
\text{Signal}(r_t) =
\begin{cases}
\text{BUY} & \text{if } r_t \geq 1.10 \\
\text{SELL} & \text{if } r_t \leq 0.90 \\
\text{HOLD} & \text{otherwise}
\end{cases}
\end{equation}
Performance is evaluated at horizons $h \in \{3, 6\}$ months using realized prices $P_{t+h}$ obtained from historical closing data with a $\pm 5$ trading-day tolerance. The three metrics of measuring stock valuation accuracy are defined as follows.

\textit{Hit Rate} measures whether the target price was reached at the end of the evaluation period. For BUY signals, a hit occurs when $P \geq \text{FVE}$; for SELL signals, when $P \leq \text{FVE}$. We report:
\begin{equation}
\text{Hit}_{\text{end}} = \mathbf{1}\left[P_{t+h} \text{ satisfies hit condition}\right]
\end{equation}

\textit{Mean Absolute Error (MAE)} captures the percentage deviation between the estimated fair value and the realized price:
\begin{equation}
\text{MAE} = \frac{1}{n} \sum_{i=1}^{n} \frac{|\text{FVE}_i - P_{t+h,i}|}{P_{t+h,i}} \times 100\%
\end{equation}

\textit{Mean Bias} measures the signed error to identify systematic optimism or conservatism:
\begin{equation}
\text{Bias} = \frac{1}{n} \sum_{i=1}^{n} \frac{\text{FVE}_i - P_{t+h,i}}{P_{t+h,i}} \times 100\%
\end{equation}
Positive bias indicates optimistic targets (FVE $>$ actual); negative bias indicates conservative targets.

\textbf{Investment Recommendation Accuracy}. Two metrics to measure investment recommendation are defined as follows.

\textit{Directional Accuracy (DA)} evaluates whether realized price movement aligns with the implied signal direction. Considering only actionable signals $s \in \mathcal{S} = \{\text{BUY}, \text{SELL}\}$:
\begin{equation}
\text{DA} = \frac{1}{|\mathcal{S}|} \sum_{s \in \mathcal{S}} \mathbf{1}\left[\text{sign}(P_{t+h} - P_t) = \text{dir}(s)\right]
\end{equation}
where $\text{dir}(\text{BUY}) = +1$ and $\text{dir}(\text{SELL}) = -1$. HOLD signals are excluded as they imply no directional prediction.

\textit{Signed Recommendation Loss (SRL)} provides a unified penalty metric that accounts for both signal direction and magnitude. Let $R_i^{\text{stock}}$ and $R_i^{\text{bench}}$ denote the stock and benchmark (S\&P 500) returns over
the horizon, with excess return $e_i = R_i^{\text{stock}} - R_i^{\text{bench}}$. Let $s_i \in \{+1, 0, -1\}$ encode BUY, HOLD, and SELL signals respectively. The SRL is defined as:
\begin{equation}
\begin{split}
\text{SRL}_i = & \; |e_i| \cdot \mathbf{1}[\text{sign}(e_i) \neq s_i] \\
            & + \lambda |e_i| \cdot \mathbf{1}[s_i = 0 \wedge |e_i| > \tau]
\end{split}
\end{equation}
The first term penalizes incorrect directional calls (BUY when stock underperformed, SELL when it outperformed). The second term penalizes HOLD recommendations when the stock moved materially beyond the threshold $\tau = 5\%$, weighted
by $\lambda = 0.33$. Lower SRL values indicate better recommendation quality.

\subsubsection{Metrics of Verifiability \& Credibility}
\label{sec:app_verify_metrics}
\textbf{Verifiability}. Let $\mathcal{C}(R)={(c_i, s_i)}_{i=1}^{N}$ denote the set of $N$ atomic claims $c_i$ extracted from $R$ along with their associated citations $s_i$. Each claim is assessed by an evidence-grounded verifier with web access, producing a binary verdict $v_i \in {0,1}$, where $v_i=1$ indicates that $c_i$ is supported by retrieved evidence consistent with $s_i$, and $v_i=0$ otherwise (unverifiable or hallucinated). We compute the factuality score as
\begin{equation}
F(R) ;=; \frac{1}{N}\sum_{i=1}^{N} v_i \times 100
\end{equation}
When $v_i=0$, we label the claim as either unverifiable or hallucinated.

\noindent\textbf{Credibility.}
\[
\mathcal{S} \;=\; \{s_1,\dots,s_M\}
\]
denote the set of all cited sources (URLs/domains) appearing in a report.

Let \(d(s)\) extract the domain of source \(s\). We define high- and low-trust domain sets
\(\mathcal{D}_{\text{hi}}\) and \(\mathcal{D}_{\text{lo}}\).
High-trust examples include \texttt{sec.gov}, \texttt{reuters.com}, \texttt{cnbc.com}, and company IR domains.
Low-trust examples include \texttt{x.com}, \texttt{investing.com}, \texttt{macrotrends.net}, and \texttt{seekingalpha.com}.

We define the investor-relations (IR) domain set as
\[
\begin{aligned}
\mathcal{D}_{\mathrm{IR}} &= \{d:\ \mathrm{IR}(d)\},\\
\mathrm{IR}(d) &\triangleq (d \text{ has prefix } \texttt{investor.}) \\
&\hspace{2.3em}\vee\ (d \text{ has prefix } \texttt{ir.}).
\end{aligned}
\]

We then partition citations into
\[
\begin{aligned}
\mathcal{S}_{\text{hi}} &= \{s\in\mathcal{S}:\ d(s)\in \mathcal{D}_{\text{hi}}\},\\
\mathcal{S}_{\text{lo}} &= \{s\in\mathcal{S}:\ d(s)\in \mathcal{D}_{\text{lo}}\}.
\end{aligned}
\]

Equivalently, indicator for source trust can be defined as:
\[
\tau(s)=
\begin{cases}
1, & d(s)\in \mathcal{D}_{\text{hi}}\\
0, & d(s)\in \mathcal{D}_{\text{lo}}\\
\end{cases}
\]

\subsubsection{LLM Judge \& Evaluation}
\label{sec:app_llm_judge}
Herein, we highlight the some of key prompt engineering considerations for better automated grading below.

\noindent\textbf{Role Contextualization}.
The prompts begin with a role statement e.g. ``you are an impartial evaluator''. This aligns with findings that LLMs perform better when contextualized with a persona or role, as it reduces instruction drift and ensures consistency in outputs \citep{wei2022cot}. 

\noindent\textbf{Dimension Definition}.
Each dimension is carefully defined in operational terms. This follows psychometric best practices where abstract qualities must be broken into observable, score-able behaviors \citep{moskal2000rubrics}. For example, "coherence" dimension can be decomposed into specific steps like "logical flow", "structural consistency" and "readability". By stating such definitions explicitly we can minimize subjective drift and improve reproducibility \citep{krippendorff2018content}.


\noindent\textbf{Grade Criteria \& Example}. Our prompts provide distinguishable criteria of different grade scale as well as relevant examples for each grade. 

\noindent\textbf{Calibration Rules}. The grade adjustment rules should be applied after the base grade. For example, in the "analytical depth", the base grade need be downgraded to "fair" if the report with "only summary information
(lists assumptions, shallow causal links" without "no quantified sensitivity analysis"

Pearson's coefficient measures the correlation as alignment between LLM judge and domain experts. Spearman’s coefficient quantifies the consistency in ranking the grading between LLM judges and domain experts.

\begin{table}[htbp]
\centering
\small
\setlength{\tabcolsep}{6pt}
\renewcommand{\arraystretch}{1.05}
\begin{adjustbox}{width=\columnwidth}
\begin{tabular}{cl|ccc}
\toprule 
\textbf{Metric} & \textbf{Report} & \textbf{GPT-5} & \textbf{Gemini 2.5} & \textbf{Sonnet 4} \\ 
                &                &                 & \textbf{Pro}        & \\
\midrule
\multirow{5}{*}{\parbox{1.5cm}{\centering Pearson Correlation ($r$)}} 
 & OpenAI  & \textbf{0.72} & \underline{0.63} & -0.03 \\
 & Gemini  & \textbf{0.63} & \underline{0.43} & 0.18 \\
 & Grok    & \textbf{0.84} & \underline{0.71} & 0.48 \\
 & Perp.   & \underline{0.55} & \textbf{0.60} & -- \\
 & Overall & \textbf{0.67} & \underline{0.61} & 0.16 \\
\midrule
\multirow{5}{*}{\parbox{1.5cm}{\centering Spearman Correlation ($\rho$)}} 
 & OpenAI  & \textbf{0.74} & \underline{0.61} & -0.03 \\
 & Gemini  & \textbf{0.63} & \underline{0.38} & 0.18 \\
 & Grok    & \textbf{0.82} & \underline{0.63} & 0.48 \\
 & Perp.   & \underline{0.56} & \textbf{0.59} & -- \\
 & Overall & \textbf{0.69} & \underline{0.60} & 0.15 \\
\bottomrule
\end{tabular}
\end{adjustbox}
\caption{Alignment between LLM judges and human annotations on qualitative evaluation of deep research reports.}
\label{tab:agreement_metrics_app}
\end{table}

\subsubsection{Prompt Templates for Evaluating Qualitative Quality}
\label{sec:app_judge_prompt}

We used LLM-as-Judge to evaluate qualitative quality of both professionals and DR agents generated research reports. Please find the prompt templates for comprehensiveness (Figure \ref{fig:prompt_Comprehensiveness}), coherence (Figure \ref{fig:prompt_coherence}), assumption (Figure \ref{fig:prompt_assumption}) and analytical depth (Figure \ref{fig:prompt_depth}) below.

\begin{figure*}[p]
\begin{slatepromptbox}{Comprehensiveness Prompt}
\tiny  
\setlength{\itemsep}{0pt}
\setlength{\parskip}{0pt}
\setlength{\topsep}{0pt}
\setlength{\partopsep}{0pt}
\setlength{\parsep}{0pt}
\texttt{<instruction>}\\[1ex]
\textbf{Role}\\
You are an impartial evaluator. Your sole task is to grade the comprehensiveness of a single equity research report.

\textbf{Definition: ``Comprehensiveness''}\\
Comprehensiveness measures whether the report fully covers the expected content areas for equity research, with proportional depth, sector-appropriate KPIs, consistent use of evidence, and minimal redundancy.

\vspace{1ex}
A comprehensive report includes:
\begin{itemize}[nosep, leftmargin=10pt, topsep=0pt, partopsep=0pt, parsep=0pt]
    \item Cover block \& contents
    \item Analyst Note / Executive Summary
    \item Business Description
    \item Business Strategy \& Outlook
    \item Bulls / Bears (or equivalent upside/downside drivers)
    \item Competitive Positioning / Economic Moat
    \item Valuation \& Profit Drivers (explicit link from drivers to valuation assumptions)
    \item Risk \& Uncertainty
    \item Capital Allocation (balance sheet, reinvestment, M\&A, buybacks/dividends)
    \item Financials Snapshot (multi-year history + forecasts, with sector-specific KPIs)
    \item ESG/Controversies (if material)
    \item Appendix/Glossary and Sources (with sources actually cited in-text)
\end{itemize}

Sector-specific KPIs should be present (e.g., ARR/NRR for SaaS, same-store sales for Retail, production volumes for Energy, NII/NIM for Banks, pipeline/trial data for Biopharma, etc.).

\textbf{What NOT to do}
\begin{itemize}[nosep, leftmargin=10pt, topsep=0pt, partopsep=0pt, parsep=0pt]
    \item Do not grade persuasiveness, style, or investment recommendation.
    \item Do not summarize the entire report. Evaluate comprehensiveness only.
    \item Do not invent facts. Use only what appears in the report.
\end{itemize}

\textbf{Evidence you must extract}\\
Identify and record the key content areas covered or missing. For each, note presence/absence and level of depth.

\textbf{Examples:}
\begin{itemize}[nosep, leftmargin=10pt, topsep=0pt, partopsep=0pt, parsep=0pt]
    \item ``Financials Snapshot includes revenue and EPS but omits sector KPIs (NIM, provisions)''
    \item ``Risk section repeats stress test results already covered in Analyst Note''
    \item ``Peer benchmarking absent''
\end{itemize}

\textbf{Hard Caps (override any other impressions)}
\begin{itemize}[nosep, leftmargin=10pt, topsep=0pt, partopsep=0pt, parsep=0pt]
    \item If $\geq$2 core sector KPIs are missing in the Financials Snapshot $\Rightarrow$ max grade = ``Fair''.
    \item If $\geq$2 material claims are uncited or based on outdated info $\Rightarrow$ max grade = ``Fair''.
    \item If valuation section does not link operating drivers to valuation outcome $\Rightarrow$ max grade = ``Fair''.
    \item If most expected sections are missing/opaque $\Rightarrow$ grade = ``Poor''.
    \item If the same talking point is repeated across $\geq$2 sections with no new depth $\Rightarrow$ downgrade one level.
\end{itemize}

\textbf{Grading Scale (choose exactly one)}
\begin{description}[nosep, leftmargin=10pt, topsep=0pt, partopsep=0pt, parsep=0pt]
    \item[Poor] Many essential sections absent or skeletal. Sector KPIs missing. Assertions uncited, outdated, or repetitive. Report feels incomplete or superficial.
    \item[Fair] Headline sections present, but details are thin. Financials Snapshot is generic or missing multiple sector KPIs. Sourcing inconsistent; limited peer/industry context. Some redundancy across sections.
    \item[Good] All major sections included with sector-appropriate KPIs. Valuation assumptions explicit and linked to drivers. Sources consistent; minimal redundancy. Risks, peers, and capital allocation addressed at reasonable depth.
    \item[Excellent] Exhaustive coverage across all sections. Every major claim tied to data/tables with sources. Valuation bridges clearly from operating drivers to fair value. Peer benchmarking and scenario/sensitivity analysis included. Zero superficial repetition.
\end{description}

\textbf{Decision Rules (apply in order)}
\begin{enumerate}[nosep, leftmargin=10pt, topsep=0pt, partopsep=0pt, parsep=0pt]
    \item If most expected sections absent/opaque $\Rightarrow$ Poor.
    \item If $\geq$2 contradictions/uncited claims/major KPI omissions $\Rightarrow$ cap at Fair.
    \item If valuation lacks explicit linkage from drivers to value $\Rightarrow$ cap at Fair.
    \item If peer/industry/regulatory context absent $\Rightarrow$ downgrade one level.
    \item If all sections, KPIs, evidence, and linkages present but no sensitivities $\Rightarrow$ Good.
    \item If comprehensive + scenario/sensitivity analysis + peer benchmarking $\Rightarrow$ Excellent.
\end{enumerate}

\texttt{</instruction>}\\[1ex]

\texttt{<output\_format>}\\
Required Output (JSON only; follow the schema exactly)

\begin{lstlisting}[basicstyle=\ttfamily\tiny, breaklines=true, columns=fullflexible]
{
  "grade": "Poor | Fair | Good | Excellent",
  "summary_reasoning": "<150-250 words covering: coverage of key sections, proportionality, evidence integration, scope vs redundancy, and gaps & impact.>",
  "content_checks": {
    "sections_present": ["Analyst Note","Business Description","Valuation","Risk","Financials Snapshot"],
    "sections_missing": ["Peer Benchmarking","Scenario Analysis"],
    "sector_kpis_present": ["Revenue","EPS"],
    "sector_kpis_missing": ["NIM","Loan Loss Provisions","CET1"]
  },
  "checks": {
    "evidence_citations_consistent": true,
    "valuation_linked_to_drivers": false,
    "peer_context_present": false,
    "redundancy_detected": true,
    "scenario_analysis_present": false
  },
  "flags": {
    "contradictions": [
      {"description": "Tax rate assumption (18%) inconsistent with industry norms", "locations": ["Valuation","Appendix"]}
    ],
    "missing_kpis": ["NIM","Provisions","CET1"],
    "uncited_claims": ["Repurchased $9.5M shares in Q1 with no source"]
  }
}
\end{lstlisting}
\texttt{</output\_format>}\\[1ex]

\texttt{<calibration\_hints>}\\
\textbf{Calibration Hints}
\begin{itemize}[nosep, leftmargin=10pt, topsep=0pt, partopsep=0pt, parsep=0pt]
    \item Reports with complete contents page but thin KPI tables, repeated stress-test mentions, and missing peer analysis $\Rightarrow$ Fair.
    \item Reports with robust KPI coverage, explicit valuation linkage, and consistent sourcing but no scenarios $\Rightarrow$ Good.
    \item Reports with exhaustive KPI coverage, peer benchmarking, and scenario/sensitivity analysis $\Rightarrow$ Excellent.
\end{itemize}
\texttt{</calibration\_hints}
\end{slatepromptbox}
\caption{Prompt template for evaluating ``comprehensiveness" metric}
\label{fig:prompt_Comprehensiveness}
\end{figure*}

\begin{figure*}[p]  
\begin{greenpromptbox}{Coherence Prompt}
\tiny  
\setlength{\itemsep}{0pt}
\setlength{\parskip}{0pt}
\setlength{\topsep}{0pt}
\setlength{\partopsep}{0pt}
\setlength{\parsep}{0pt}
\texttt{<instruction>}\\[1ex]
\textbf{Role}

You are an impartial grader. Evaluate the coherence of a single equity research report (text + any included tables) using only the content provided. Do not add outside facts.

\textbf{Coherence} = how well the report reads as one connected argument: clear language, smooth local transitions, explicit cross-references among sections, and no internal contradictions among text, tables, assumptions, dates, and cited facts.

\textbf{What To Read}
\begin{itemize}[nosep, leftmargin=10pt, topsep=0pt, partopsep=0pt, parsep=0pt]
    \item All narrative sections (e.g., Analyst Note, Strategy/Outlook, Bulls/Bears, Moat, Valuation/Assumptions, Risks, Capital Allocation, ESG, Appendix, Sources).
    \item All tables/figures embedded in the report.
\end{itemize}

\textbf{Step-by-Step Procedure}
\begin{enumerate}[nosep, leftmargin=10pt, topsep=0pt, partopsep=0pt, parsep=0pt]
    \item \textbf{Scan \& Map}: Identify section headers and the overall narrative arc (intro $\rightarrow$ analysis $\rightarrow$ conclusion). Note defined terms/jargon and whether they are used consistently.
    \item \textbf{Flow Check (Within \& Across Sections):}
    \begin{itemize}[nosep, leftmargin=10pt, topsep=0pt, partopsep=0pt, parsep=0pt]
        \item \textbf{Within}: Do paragraphs transition logically, stating why the next idea follows?
        \item \textbf{Across}: Do sections reference each other (e.g., assumptions $\leftrightarrow$ valuation $\leftrightarrow$ risks) rather than read like isolated lists?
    \end{itemize}[nosep, leftmargin=10pt, topsep=0pt, partopsep=0pt, parsep=0pt]
    \item \textbf{Consistency Check}: Find contradictions or unresolved tensions among:
    \begin{itemize}[nosep, leftmargin=10pt, topsep=0pt, partopsep=0pt, parsep=0pt]
        \item Narrative vs tables (e.g., ``capex flat'' vs large recurring tech spend in the same period).
        \item Assumptions vs Bulls/Bears vs base-case valuation (e.g., ``NII steadily rising'' vs rate-cut sensitivity).
        \item Temporal claims vs dates/citations (e.g., calling an older event ``recent'').
        \item Terminology use (e.g., NIM vs NII misapplied).
    \end{itemize}[nosep, leftmargin=10pt, topsep=0pt, partopsep=0pt, parsep=0pt]
    \item \textbf{Citations Alignment}: If sources are cited, do they support the claim as written (at least at a basic face-value level)? If not, note the mismatch.
    \item Apply the Rubric \& Guardrails (below) and determine the final grade.
    \item Write Reasoning (180-260 words): Explain the key drivers of your grade with direct references to sections/lines where possible.
    \item Select 2-3 Evidence Bullets: Quote or concisely paraphrase the most decisive strengths/weaknesses with a section/page cue if available.
\end{enumerate}

\textbf{Grading Scale (choose exactly one)}
\begin{description}[nosep, leftmargin=10pt, topsep=0pt, partopsep=0pt, parsep=0pt]
    \item[poor] Frequent abrupt jumps; list-like paragraphs; key terms undefined or inconsistently used. Multiple unresolved contradictions (narrative $\leftrightarrow$ tables/assumptions) or time-inaccurate ``recency'' claims. Citations/figures not aligned with the stated point. Reader must assemble the argument themselves.
    \item[fair] Overall structure exists, but within-section flow is choppy; limited transitions. Some contradictions or unexplained tensions remain; terms mostly clear but occasionally inconsistent. Understandable with effort, but internally noisy.
    \item[good] Clear language and mostly smooth transitions; sections reinforce each other. No material contradictions; dates/sources align. Minor rough edges (e.g., a thin transition or lightly supported claim) are acceptable.
    \item[excellent] Seamless narrative with explicit connective tissue (``Because X in Q1, therefore Y in margins''). Assumptions, tables, and narrative perfectly align; terminology consistent; temporal claims precise. Proactively surfaces and resolves potential contradictions.
\end{description}

\textbf{Guardrails (Auto-Downgrades)}
\begin{itemize}[nosep, leftmargin=10pt, topsep=0pt, partopsep=0pt, parsep=0pt]
    \item Hard cap = fair if you find any unaddressed contradiction among assumptions, tables, and narrative.
    \item Hard cap = fair if any ``recent/now'' claims are time-inaccurate relative to dates provided.
    \item Minus one level if Bulls/Bears are not reconciled with the base-case valuation narrative.
    \item Minus one level if $\geq$2 sections read like bullet dumps (no transitions).
\end{itemize}

\texttt{</instruction>}\\[1ex]

\texttt{<output\_format>}\\
Output Format (return only this JSON object)

\begin{lstlisting}[basicstyle=\ttfamily\tiny, breaklines=true, columns=fullflexible]
{
  "grade": "poor | fair | good | excellent",
  "reasoning": "<180-260 words explaining the grade, touching on clarity, flow, structural coherence, consistency, and temporal/source alignment>",
  "evidence": [
    "<brief quote or paraphrase + section/page cue>",
    "<brief quote or paraphrase + section/page cue>",
    "<optional third bullet>"
  ],
  "flags": {
    "contradictions_found": true,
    "temporal_inaccuracy_found": true,
    "bulls_bears_unreconciled": true,
    "list_like_sections_count": 0,
    "auto_downgrade_applied": true
  }
}
\end{lstlisting}
\texttt{</output\_format>}\\[1ex]

\texttt{<calibration\_hints>}\\
\textbf{Calibration Hints (so the weak example scores Poor/Fair; strong reports score Good/Excellent)}
\begin{itemize}[nosep, leftmargin=10pt, topsep=0pt, partopsep=0pt, parsep=0pt]
    \item If you detect contradictions like ``capex/share count flat'' alongside repeated, large tech spend claims, or ``NII steadily rising'' while Bears emphasize rate-cut sensitivity without reconciliation, set contradictions\_found = true and cap at fair.
    \item If a ``recent'' event is actually older within the report's own dates/sources, set temporal\_inaccuracy\_found = true and cap at fair.
    \item If the narrative is polished but jumps topic (earnings $\rightarrow$ tech $\rightarrow$ stress tests) with thin transitions, count those sections toward list\_like\_sections\_count and consider a fair grade unless other strengths clearly outweigh.
\end{itemize}
\texttt{</calibration\_hints}
\end{greenpromptbox}
\caption{Prompt template for evaluating ``coherence" metric}
\label{fig:prompt_coherence}
\end{figure*}

\begin{figure*}[p]
\begin{burgundypromptbox}{Assumptions Prompt}
\tiny  
\setlength{\itemsep}{0pt}
\setlength{\parskip}{0pt}
\setlength{\topsep}{0pt}
\setlength{\partopsep}{0pt}
\setlength{\parsep}{0pt}
\texttt{<instruction>}\\[1ex]
\textbf{Role}

You are an impartial evaluator. Your sole task is to grade the quality of assumptions in a single equity research report.

\textbf{Definition: ``Quality of Assumptions''}

Quality of assumptions measures whether the report's key inputs that drive forecasts, valuations, and conclusions are:
\begin{itemize}[nosep, leftmargin=10pt, topsep=0pt, partopsep=0pt, parsep=0pt]
    \item \textbf{Explicit} (clearly stated near the forecasts they inform),
    \item \textbf{Justified} (supported by multi-year history, peer/industry benchmarks, or cited sources),
    \item \textbf{Specific} (magnitudes, units, time horizons, and causal drivers),
    \item \textbf{Consistent} (internally coherent with the report's own tables and conclusions), and
    \item \textbf{Stress-tested} (quantified ranges or scenarios for material drivers, not only a single point).
\end{itemize}

\textbf{What NOT to do}
\begin{itemize}[nosep, leftmargin=10pt, topsep=0pt, partopsep=0pt, parsep=0pt]
    \item Do not grade writing quality, persuasion, or investment merit.
    \item Do not summarize the whole report. Evaluate assumptions only.
    \item Do not invent facts. Use only what appears in the report.
\end{itemize}

\textbf{Evidence you must extract}

Identify and quote (verbatim, short excerpts) the top 3-8 assumptions that materially drive the model (e.g., revenue/NII growth, margins/efficiency ratio, credit losses, capex, tax rate, WACC/COE, terminal growth, share count). For each, record where it appears (section header and page if available).

\textbf{Hard Caps (override any other impressions)}
\begin{itemize}[nosep, leftmargin=10pt, topsep=0pt, partopsep=0pt, parsep=0pt]
    \item No sensitivity/scenario analysis on material drivers $\Rightarrow$ max grade = ``Fair''.
    \item Any unresolved numeric contradiction between text and tables/figures $\Rightarrow$ max grade = ``Fair''.
    \item Assumptions mostly unstated/opaque $\Rightarrow$ ``Poor''.
    \item Use of figures that clearly conflict with recent history/peers without justification $\Rightarrow$ max grade = ``Fair''.
    \item Two or more distinct contradictions (or one contradiction plus no sensitivities) $\Rightarrow$ ``Poor''.
\end{itemize}

\textbf{Grading Scale (choose exactly one)}
\begin{description}[nosep, leftmargin=10pt, topsep=0pt, partopsep=0pt, parsep=0pt]
    \item[Poor] Assumptions largely unstated/opaque or scattered without context. Little/no justification; conflicts with facts or the report's own tables. Vague magnitudes (no units, horizons, or drivers). No sensitivities/scenarios; outputs feel black-box. Often includes internal contradictions (e.g., ``flat share count'' while committing to heavy ongoing tech spend).
    \item[Fair] Key assumptions are listed but weakly justified; some are selective/hand-wavy. Partial specificity (some units/timeframes) with minor inconsistencies. Limited/qualitative sensitivity (mentions risk without quantified ranges).
    \item[Good] Assumptions are explicit and proximal to the forecasts they drive; units, horizons, and drivers are clear. Justified with history/peers/sources; numbers reconcile to tables and narrative. Quantified sensitivity for at least 2-3 material drivers (e.g., rates/NIM, credit cost, margin).
    \item[Excellent] Assumptions are comprehensive, traceable, auditable across narrative $\leftrightarrow$ tables $\leftrightarrow$ conclusions. Evidence-rich (multi-year history, peer comps, and/or cited sources) for each major driver. Robust, quantified sensitivities/scenarios for all material levers with valuation/target impact. Zero contradictions; figures, units, and time windows reconcile throughout.
\end{description}

\textbf{Decision Rules (apply in order)}
\begin{enumerate}[nosep, leftmargin=10pt, topsep=0pt, partopsep=0pt, parsep=0pt]
    \item If assumptions are mostly unstated/opaque $\Rightarrow$ Poor.
    \item Else, count contradictions (text vs table; claim vs clearly implied historical fact in the report).
    \item $\geq$2 contradictions $\Rightarrow$ Poor.
    \item 1 contradiction $\Rightarrow$ cap at Fair (unless fully reconciled in text).
    \item Check sensitivities/scenarios on material drivers.
    \item None/qualitative only $\Rightarrow$ cap at Fair.
    \item Check justification (history, peers, sources) and specificity (units/horizons/drivers).
    \item Weak/partial $\Rightarrow$ Fair.
    \item Solid with some quantified sensitivities $\Rightarrow$ Good.
    \item Comprehensive + robust sensitivities on all material levers $\Rightarrow$ Excellent.
\end{enumerate}

\texttt{</instruction>}\\[1ex]

\texttt{<output\_format>}\\
Required Output (JSON only; follow the schema exactly)

\begin{lstlisting}[basicstyle=\ttfamily\tiny, breaklines=true, columns=fullflexible]
{
  "grade": "Poor | Fair | Good | Excellent",
  "summary_reasoning": "<150-250 words focusing on the five pillars: explicitness, justification, specificity, consistency, sensitivity. No fluff.>",
  "assumptions_extracted": [
    {
      "quote": "<verbatim short excerpt>",
      "location": {"section": "<header>", "page": "<number or 'unknown'>"},
      "driver_type": "<e.g., revenue_growth | NII/NIM | margin/efficiency | credit_cost | tax_rate | capex | share_count | WACC/COE | terminal_growth | other>"
    }
  ],
  "checks": {
    "explicitness": true,
    "justification_with_evidence": "none | weak | partial | solid | comprehensive",
    "specificity_units_horizon": "none | weak | partial | solid | comprehensive",
    "internal_consistency": "clean | minor_issues | contradiction_found",
    "sensitivities_present": false,
    "sensitivities_quality": "none | qualitative_only | partial_quant | robust_quant",
    "material_drivers_covered": ["NII/NIM","margin/efficiency","WACC/COE","terminal_growth","share_count"]
  },
  "flags": {
    "contradictions": [
      {"description": "<what conflicts with what>", "locations": ["<where A>", "<where B>"]}
    ],
    "missing_or_opaque_assumptions": ["<driver types missing>"],
    "unjustified_parameters": ["<e.g., WACC 9% without source>"]
  }
}
\end{lstlisting}
\texttt{</output\_format>}\\[1ex]

\texttt{<calibration\_hints>}\\
\textbf{Calibration Hints (to guide borderline calls)}
\begin{itemize}[nosep, leftmargin=10pt, topsep=0pt, partopsep=0pt, parsep=0pt]
    \item Reports that list assumptions but include incorrect/unsupported figures, internal contradictions, and no quantified sensitivities should end up Poor/Fair (the ``leniency trap'' is to over-reward the list itself—don't).
    \item Reports that state and justify inputs (e.g., efficiency ratio path, expense CAGR, credit losses, NII/NIM bands, WACC tied to market yields) and offer quantified scenarios for major levers should be Good/Excellent.
\end{itemize}

\textbf{Final constraints}
\begin{itemize}[nosep, leftmargin=10pt, topsep=0pt, partopsep=0pt, parsep=0pt]
    \item Keep the summary\_reasoning within 150-250 words.
    \item Return valid JSON matching the schema.
    \item If a required field is unavailable, use ``unknown'' or an empty array as appropriate—do not invent content.
\end{itemize}
\texttt{</calibration\_hints}
\end{burgundypromptbox}
\caption{Prompt template for evaluating ``assumptions" metric}
\label{fig:prompt_assumption}
\end{figure*}

\begin{figure*}[p]  
\begin{navypromptbox}{Analytical Depth Prompt}
\tiny  
\setlength{\itemsep}{0pt}
\setlength{\parskip}{0pt}
\setlength{\topsep}{0pt}
\setlength{\partopsep}{0pt}
\setlength{\parsep}{0pt}
\texttt{<instruction>}\\[1ex]
\textbf{Role}\\
You are an impartial evaluator. Your sole task is to grade the analytical depth of a single equity research report.

\textbf{Definition: ``Analytical Depth''}\\
Analytical depth measures whether the report goes beyond description to provide causal reasoning, explicit assumptions, quantified forecasts, counterpoints/uncertainty, and decision-relevant implications.

A report with analytical depth includes:
\begin{itemize}[nosep, leftmargin=10pt, topsep=0pt, partopsep=0pt, parsep=0pt]
    \item \textbf{Causal Explanation}: why results occur, not just what happened
    \item \textbf{Inference Quality}: forecasts or hypotheses tied to explicit drivers
    \item \textbf{Use of Data}: quantified assumptions, benchmarks, sanity checks, clear distinction between correlation vs causation
    \item \textbf{Counterpoints \& Uncertainty}: scenarios, sensitivities, ranges, limitations, alternative explanations
    \item \textbf{Actionable Implications}: clear, conditional takeaways relevant for investor decisions
\end{itemize}

\textbf{What NOT to do}
\begin{itemize}[nosep, leftmargin=10pt, topsep=0pt, partopsep=0pt, parsep=0pt]
    \item Do not grade persuasiveness, writing style, or investment recommendation.
    \item Do not summarize the entire report. Evaluate analytical depth only.
    \item Do not invent facts. Use only what appears in the report.
\end{itemize}

\textbf{Evidence you must extract}\\
Identify and record the strongest and weakest examples of reasoning. Quote 3--6 short verbatim excerpts (\textless 10 words) from the report that illustrate causal reasoning, assumptions, sensitivities, or lack thereof.

\textbf{Examples:}
\begin{itemize}[nosep, leftmargin=10pt, topsep=0pt, partopsep=0pt, parsep=0pt]
    \item ``NII was \$14.6B, driven by higher yielding assets'' (causal)
    \item ``Operating costs grow slower than revenue due to scale'' (causal mechanism)
    \item ``If Fed cuts rates, NIM could compress'' (counterpoint)
    \item ``WACC $\sim$9\%, terminal growth $\sim$2.5\%'' (assumption, but no sensitivity)
    \item ``Projected steadily rising NII'' (unsupported generalization)
\end{itemize}

\textbf{Hard Caps (override any other impressions)}
\begin{itemize}[nosep, leftmargin=10pt, topsep=0pt, partopsep=0pt, parsep=0pt]
    \item If no clear mechanisms or assumptions are given -- max grade = ``Poor''.
    \item If assumptions are listed but not benchmarked or stress-tested -- max grade = ``Fair''.
    \item If report contains explicit assumptions and causal reasoning but no quantified sensitivity/scenario -- max grade = ``Good''.
    \item Only reports with mechanism-rich reasoning, quantified scenarios/ranges, and actionable implications -- ``Excellent''.
\end{itemize}

\textbf{Grading Scale (choose exactly one)}
\begin{description}[nosep, leftmargin=10pt, topsep=0pt, partopsep=0pt, parsep=0pt]
    \item[Poor] Mostly descriptive. No mechanisms, little or no assumptions, no quantified sensitivity.\\
    Example: Lists ``EPS grows'' without linking to costs, margins, or rates.
    \item[Fair] Some mechanisms/assumptions, but shallow, unsupported, or generic. Limited quantification. Counterpoints mentioned but not explored. Implications vague.\\
    Example: ``NII up on higher yields'' but no deposit beta or rate sensitivity.
    \item[Good] Clear mechanisms tied to forecasts and valuation. Explicit assumptions quantified and benchmarked. Some scenario/counterpoint analysis but not extensive.\\
    Example: ``100 bp cut lowers NIM $\sim$20 bps, EPS down $\sim$5\%.'' 
    \item[Excellent] Comprehensive causal reasoning with explicit, benchmarked assumptions. Multiple quantified scenarios or ranges. Strong treatment of uncertainty. Clear, conditional implications.\\
    Example: ``Base case ROTE 14\% vs COE 9.5\%; if loan losses +50 bps, EPS --8\% and FV drops \$50--\$44.''
\end{description}

\textbf{Decision Rules (apply in order)}
\begin{enumerate}[nosep, leftmargin=10pt, topsep=0pt, partopsep=0pt, parsep=0pt]
    \item If no mechanisms/assumptions -- Poor.
    \item If mechanisms/assumptions exist but no benchmarks/sensitivity -- Fair.
    \item If mechanisms and assumptions are benchmarked but no scenario analysis -- Good.
    \item If all of the above + scenario/sensitivity analysis and decision-relevant implications -- Excellent.
\end{enumerate}

\texttt{</instruction>}\\

\texttt{<output\_format>}\\
Required Output (JSON only; follow the schema exactly)
\begin{lstlisting}[basicstyle=\ttfamily\tiny, breaklines=true, columns=fullflexible, aboveskip=1pt, belowskip=1pt]
{
  "grade": "Poor | Fair | Good | Excellent",
  "summary_reasoning": "lt 150 to 250 words covering causal explanation, inference quality, data use, counterpoints/uncertainty, and actionable implications. Include 3 to 6 short verbatim excerpts.",
  "checks": {
    "causal_explanation_present": true,
    "assumptions_explicit": true,
    "assumptions_benchmarked": false,
    "quantification_used": true,
    "sensitivity_or_scenarios": false,
    "actionable_implications_present": true
  },
  "flags": {
    "missing_mechanisms": ["balance sheet repositioning - yield calibration not explained"],
    "unsupported_assumptions": ["terminal growth 2.5% with no industry benchmark"],
    "lack_of_sensitivity": ["NIM assumption fixed at 3.3% without range"]
  }
}
\end{lstlisting}
\texttt{</output\_format>}\\

\texttt{<calibration\_hints>}\\
\textbf{Calibration Hints}
\begin{itemize}[nosep, leftmargin=10pt, topsep=0pt, partopsep=0pt, parsep=0pt]
    \item Reports with only summary information (lists assumptions, shallow causal links, no quantified sensitivity) $\Rightarrow$ Fair.
    \item Reports with explicit assumptions, benchmarks, and some causal reasoning but no scenarios $\Rightarrow$ Good.
    \item Reports with explicit assumptions, benchmarks, quantified scenarios, and decision-relevant implications $\Rightarrow$ Excellent.
\end{itemize}
\texttt{</calibration\_hints}
\end{navypromptbox}
\caption{Prompt template for evaluating ``analytical depth" metric}
\label{fig:prompt_depth}
\end{figure*}

\subsection{Supplemental Evaluation Results}
\label{sec:app_supp_eval}
We included the additional evaluation results, which cannot fit into the main paper, in this section of appendix.

\subsubsection{Evaluations of Deep Research Agents on Firm B}
\label{sec:app_firmb_eval}

Due to the limited space, the main paper only presented the evaluation results of deep research agents on Firm A dataset. Herein, we also include the evaluation results on Firm B dataset which used the same list of companies and fiscal quarters as Firm A dataset. Firm B is an independent provider of market intelligence to commodity and equity markets. It is less well known among global financial institutions. Similar to Firm A, we first used LLMs to extract rubrics from an example professional reports from Firm B and then create an input prompt to deep research agents based on the rubrics.

\noindent\textbf{Qualitative Evaluation}. Consistent with the observations from Firm A dataset, professionals still yield higher quality in all four qualitative dimensions than frontier deep research agents  except assumption. Gemini DR is still the strongest performing DR agents.   

\begin{table}[htbp]
\centering
\small
\setlength{\tabcolsep}{6pt}
\renewcommand{\arraystretch}{1.05}
\begin{adjustbox}{width=\columnwidth}
\begin{tabular}{l|c|c|c|c|c}
\toprule
\textbf{Report} & \textbf{Comp.} & \textbf{Assum.} & \textbf{Cohe.} & \textbf{Dep.} & \textbf{Overall} \\
\hline
\multicolumn{6}{c}{Deep Research Agents} \\
\hline
OAI        
& 1.68 
& 1.20 
& 1.32 
& 2.08 
& 1.57 \\

Gemini     
& \uline{1.80} 
& \textbf{1.80} 
& 1.40 
& 2.12 
& \uline{1.78} \\

Grok       
& 1.58 
& 1.32 
& 1.20 
& 2.08 
& 1.55 \\

Perplexity 
& 1.76 
& 1.28 
& \uline{1.44} 
& \uline{2.24} 
& 1.68 \\

\hline
\multicolumn{6}{c}{Professional Analysts} \\
\hline
Firm B 
& \textbf{2.00} 
& \uline{1.64} 
& \textbf{1.92} 
& \textbf{2.28} 
& \textbf{1.96} \\

\bottomrule
\end{tabular}
\end{adjustbox}
\caption{Model performance statistics showing mean scores across metrics (scale: 1=Poor, 2=Fair, 3=Good, 4=Excellent). Bold indicates best and underlined indicates second-best within each metric. Reports were generated using a prompt template based on Firm B dataset.}
\label{tab:model_performance_firm_b}
\end{table}

\noindent\textbf{Quantitative Evaluation}. Professionals reports from Firm B only contain financial projections of revenue and EPS. Therefore, we benchmarked these two metrics against DR reports as well. Results in Table \ref{tab:forecast_accuracy_smape_revenue_eps} indicate professionals did significant better job on predicting EPS than DR agents, but less accurate on forecasting revenue. 

\begin{table}[htbp]
\centering
\small
\setlength{\tabcolsep}{6pt}
\renewcommand{\arraystretch}{1.10}
\begin{adjustbox}{width=\columnwidth}
\begin{tabular}{l|c|c|c|c|c}
\toprule
\textbf{Metric}
& \textbf{OpenAI}
& \textbf{Gemini}
& \textbf{Grok}
& \textbf{Perp.}
& \textbf{Firm B} \\
\midrule

Revenue
& 20.35
& \textbf{16.87}
& \uline{17.33}
& 23.64
& 21.74 \\

EPS
& 34.61
& 23.75
& 16.24
& \uline{10.28}
& \textbf{3.66} \\

\bottomrule
\end{tabular}
\end{adjustbox}

\caption{
Forecast accuracy measured by SMAPE (\%).
Lower values indicate better performance.
Bold indicates best result per metric; underlining indicates second-best.}
\label{tab:forecast_accuracy_smape_revenue_eps}
\end{table}

In valuating the target stock price for next 3-6 months, professionals overall can make a bit better prediction of future than best DR agent, OpenAI. Although OpenAI is more accurate within the 3-month period, professionals have the highest hit rate and lowest error against actual stock prices within the 6-month period.

\begin{table}[htbp]
\centering
\scriptsize
\setlength{\tabcolsep}{4pt}
\renewcommand{\arraystretch}{1.1}
\begin{adjustbox}{width=0.9\columnwidth}
\begin{tabular}{l|c|ccc}
\toprule
\multirow{2}{*}{\textbf{Report}} & \multirow{2}{*}{\textbf{Horizon}} & \textbf{Hit Rate} & \textbf{MAE} & \textbf{Mean Bias} \\
                &                  &   (\%)            &  (\%)        & (\%) \\
\hline
\multicolumn{5}{c}{Deep Research Agents} \\
\hline
\multirow{2}{*}{OAI}
& 3m & \textbf{31.8} & \textbf{16.3} & \uline{+10.7} \\
& 6m & 52.4 & 17.3 & +4.1 \\
\hline
\multirow{2}{*}{Gemini}
& 3m & 22.3 & 19.8 & +19.8 \\
& 6m & 47.6 & \uline{16.0} & +13.1 \\
\hline
\multirow{2}{*}{Perp.}
& 3m & 20.8 & 18.1 & +16.0 \\
& 6m & 43.5 & 17.5 & +9.3 \\
\hline
\multirow{2}{*}{Grok}
& 3m & 23.1 & 26.3 & \textbf{+7.7} \\
& 6m & \uline{53.8} & 25.3 & \textbf{-0.0} \\
\hline
\multicolumn{5}{c}{Professional Analysts} \\
\hline
\multirow{2}{*}{Firm B}
& 3m & \uline{27.8} & \uline{17.4} & +10.8 \\
& 6m & \textbf{55.6} & \textbf{12.9} & \uline{-0.1} \\
\bottomrule
\end{tabular}
\end{adjustbox}
\caption{Target price metrics by horizon (3m and 6m only). Hit rate: higher is better. MAE: lower is better. Mean bias ranking uses absolute magnitude (closer to zero is better) while retaining sign. Bold indicates best and underlined indicates second-best within each horizon and metric.}
\label{tab:target_price_metrics_3m6m}
\end{table}

In terms of stock recommendations of ``buy", ``sell" or ``hold", professionals were also more accurate in comparison with the actual stock movement within 6 months, even if OpenAI was marginally better during the first 3 months.    

\begin{table}[htbp]
\centering
\setlength{\tabcolsep}{4pt}
\renewcommand{\arraystretch}{1.1}
\tiny
\begin{adjustbox}{width=0.8\columnwidth}
\begin{tabular}{l|c|cc}
\toprule
\textbf{Report} & \textbf{Horizon} & \textbf{Dir. Acc. (\%)} & \textbf{SRL} \\
\hline
\multicolumn{4}{c}{Deep Research Agents} \\
\hline
\multirow{2}{*}{OAI}
& 3m & \textbf{81.8} & \textbf{0.052} \\
& 6m & 71.4 & 0.071 \\
\hline
\multirow{2}{*}{Gemini}
& 3m & 50.0 & 0.062 \\
& 6m & \uline{75.0} & 0.074 \\
\hline
\multirow{2}{*}{Perp.}
& 3m & 75.0 & 0.062 \\
& 6m & 73.9 & \uline{0.068} \\
\hline
\multirow{2}{*}{Grok}
& 3m & 53.8 & 0.079 \\
& 6m & 61.5 & 0.076 \\
\hline
\multicolumn{4}{c}{Professional Analysts} \\
\hline
\multirow{2}{*}{Firm B}
& 3m & \uline{77.8} & \uline{0.060} \\
& 6m & \textbf{83.3} & \textbf{0.050} \\
\bottomrule
\end{tabular}
\end{adjustbox}
\caption{Stock recommendation metrics by horizon (3m and 6m only). Directional accuracy: higher is better. SRL: lower is better. Bold indicates best and underlined indicates second-best within each horizon and metric.}
\label{tab:stock_rec_metrics_3m6m}
\end{table}

Overall, there are mixed results from quantitative evaluation. Professional analysts from Firm B made more accurate predictions and recommendations in certain occasions, but not always. Evaluations on firm A in the main paper show the similar trends. 

\noindent\textbf{Verifiability \& Credibility Evaluation}. Here we discuss verifiability results of Firm B, and credibility results for both Firm A and Firm B.  

\begin{table}[htbp]
\centering
\small
\setlength{\tabcolsep}{6pt}
\renewcommand{\arraystretch}{1.05}
\begin{adjustbox}{width=\columnwidth}
\begin{tabular}{c|cccc}
\toprule 
 \textbf{Metrics} & \textbf{OpenAI} & \textbf{Gemini} & \textbf{Grok} & \textbf{Perp.} \\
 \midrule
 \# claims        &  27.8   & 27.1   & 20.4   & 29.9 \\
 \# num. claims   &  25.4   & 24.1   & 18.7   & 26.9 \\
 \# desc. claims  &  3.1    & 3.9    & 2.2    & 3.2  \\
 \hline
 \hline 
overall $F(R)$    &  \textbf{83.0\%} & 45.0\% & 40.6\% & \uline{81.0\%} \\
num. $F(R)$       &  \textbf{82.5\%} & 42.3\% & 39.0\% & \uline{81.4\%} \\
desc. $F(R)$      &  \textbf{89.6\%} & 53.9\% & 50.7\% & \uline{77.6\%} \\
overall $H(R)$    &  \textbf{10.8\%} & \uline{11.7\%} & 45.7\% & 17.2\% \\
overall $NV(R)$   &  \uline{6.1\%}   & 43.3\% & 13.7\% & \textbf{1.7\%} \\
\bottomrule
\end{tabular}
\end{adjustbox}
\caption{Verifiability analysis of DR agent-generated reports where the generated reports follow the structure
of Firm B. Bold indicates best and underlined indicates second-best within each metric.}
\label{tab:hallucination_summary_arg}
\end{table}

\noindent\textbf{Verifiability.}
For Firm B, we can find a significant reduction in $F(R)$ for Grok and Gemini. However, the hallucination rate for Gemini is still lower with respect to Grok's hallucination rate. However, Gemini has produced a noticeable number of unverifiable citations.

\noindent\textbf{Credibility.}
 Table \ref{tab:trust_citation_rates} shows the credibility results. It can be noticed that for both types of reports (Firm A and Firm B), OpenAI and Perplexity deep research would provide higher percentage of high-trust sources. While for Gemini and Grok, the percentage of high-trust sources is significantly lower.

\begin{table}[t]
\centering
\caption{High- vs. Low-trust web citation percentages by DR agents for Firm A and Firm B.}
\label{tab:trust_citation_rates}
\setlength{\tabcolsep}{6pt}
\renewcommand{\arraystretch}{1.00}
\begin{adjustbox}{width=0.8\columnwidth}
\begin{tabular}{lcccc}
\hline
\textbf{Model} & \multicolumn{2}{c}{\textbf{Firm A (\%)}} & \multicolumn{2}{c}{\textbf{Firm B (\%)}} \\
\cline{2-3}\cline{4-5}
 & \textbf{High} & \textbf{Low} & \textbf{High} & \textbf{Low} \\
\hline
Gemini      & 61.00 & 39.00 & 59.40 & 40.60 \\
Grok  & 54.81 & 45.19 & 65.29 & 34.71 \\
OAI         & 88.21 & 11.79 & 83.10 & 16.90 \\
Perplexity  & 74.57 & 25.43 & 77.22 & 22.78 \\
\hline
\end{tabular}
\end{adjustbox}
\end{table}

\subsubsection{Evaluations of LLMs with Web Search}
\label{sec:app_llm_web_eval}

Besides deep research agents, we also experimented with the non-AI agent solution which only leverages LLM + web search to generate research reports. We mainly used two OpenAI (gpt4o-mini and gpt4o) and two Perplexity (sonar-pro and sonar-reasoning-pro) LLMs to compare with the corresponding deep research agents from the same vendors for this evaluation.

\noindent\textbf{Qualitative Evaluation}. Tables \ref{tab:model_performance_llm_web_firm_a} and \ref{tab:model_performance_llm_web_firm_b} present qualitative metrics on Firm A and B datasets respectively. Deep research agents consistently outperform LLMs with web search, with Gemini achieving the highest overall scores on both datasets (2.30 on Firm A, 1.78 on Firm B). The performance gap is more pronounced on Firm A, where Gemini exceeds the best LLM (sonar-pro, 1.70) by 0.60 points, compared to 0.24 points on Firm B. Among LLMs with web search, sonar-pro leads on both datasets, while gemini-2.5-flash shows mixed performance (1.59 on Firm A, 1.25 on Firm B).

\begin{table}[htbp]
\centering
\small
\setlength{\tabcolsep}{6pt}
\renewcommand{\arraystretch}{1.05}
\begin{adjustbox}{width=\columnwidth}
\begin{tabular}{l|ccccc}
\toprule
\textbf{Report} & \textbf{Comp.} & \textbf{Assum.} & \textbf{Cohe.} & \textbf{Dep.} & \textbf{Overall} \\
\hline
\multicolumn{6}{c}{LLMs + Web Search} \\
\hline
gpt4o-mini        
& 1.20 
& 1.40 
& 1.10 
& 2.00
& 1.42 \\

gpt4o     
& 1.20 
& 1.40 
& 1.40 
& 2.00 
& 1.50 \\

sonar-pro       
& 1.70 
& 1.40 
& 1.20 
& 2.50 
& 1.70 \\

sonar-reason-pro 
& 1.50 
& 1.22 
& 1.50 
& 1.90 
& 1.54 \\

gemini-2.5-flash 
& 1.68 
& 1.36 
& 1.29 
& 2.00 
& 1.59 \\
\hline
\multicolumn{6}{c}{Deep Research Agents} \\
\hline
OAI        
& \uline{2.00} 
& 1.45 
& \uline{2.10} 
& \uline{2.60} 
& 2.02  \\ 

Gemini     
& \textbf{2.09} 
& \textbf{2.00} 
& \textbf{2.20} 
& \textbf{3.00} 
& \textbf{2.30} \\

Perplexity 
& \uline{2.00}       
& \uline{1.90}  
& \textbf{2.20}  
& 2.40   
& \uline{2.12}  \\

\bottomrule
\end{tabular}
\end{adjustbox}
\caption{Model performance statistics showing mean scores across metrics (scale: 1=Poor, 2=Fair, 3=Good, 4=Excellent). Reports were generated using a prompt template based on Firm A dataset. Bold indicates best and underlined indicates second-best within each metric.}
\label{tab:model_performance_llm_web_firm_a}
\end{table}

\begin{table}[htbp]
\centering
\small
\setlength{\tabcolsep}{6pt}
\renewcommand{\arraystretch}{1.05}
\begin{adjustbox}{width=\columnwidth}
\begin{tabular}{l|ccccc}
\toprule
\textbf{Report} & \textbf{Comp.} & \textbf{Assum.} & \textbf{Cohe.} & \textbf{Dep.} & \textbf{Overall} \\
\hline
\multicolumn{6}{c}{LLMs + Web Search} \\
\hline
gpt4o-mini        
& 1.28 
& 1.00 
& 1.16 
& 1.84 
& 1.32 \\

gpt4o     
& 1.32 
& 1.08 
& 1.24 
& 1.96 
& 1.40 \\

sonar-pro       
& 1.60 
& 1.24 
& 1.24 
& 2.08 
& 1.54 \\

sonar-reason-pro 
& 1.40 
& \uline{1.28} 
& \uline{1.40} 
& 1.80 
& 1.47 \\

gemini-2.5-flash 
& 1.24 
& 1.08 
& 1.04 
& 1.64 
& 1.25 \\
\hline
\multicolumn{6}{c}{Deep Research Agents} \\
\hline
OAI        
& 1.68  
& 1.20 
& 1.32 
& 2.08 
& 1.57 \\

Gemini     
& \textbf{1.80} 
& \textbf{1.80} 
& \uline{1.40} 
& \uline{2.12} 
& \textbf{1.78} \\

Perplexity 
& \uline{1.76} 
& \uline{1.28} 
& \textbf{1.44} 
& \textbf{2.24} 
& \uline{1.68}  \\

\bottomrule
\end{tabular}
\end{adjustbox}
\caption{Model performance statistics showing mean scores across metrics (scale: 1=Poor, 2=Fair, 3=Good, 4=Excellent) based on Firm B prompt style. Bold indicates best and underlined indicates second-best within each metric.}
\label{tab:model_performance_llm_web_firm_b}
\end{table}

\noindent\textbf{Quantitative Evaluation - Financial Projections}. Table \ref{tab:llm_web_fin_forecast_smape_a} and \ref{tab:llm_web_fin_forecast_smape_b} showed the performance of financial projections on Firm A and B datasets respectively. Although some LLMs with web search were modestly more accurate on revenue and EPS forecasts, deep research agents achieved considerably higher performance on the remaining financial metrics.

\begin{table}[htbp]
\centering
\small
\setlength{\tabcolsep}{5pt}
\renewcommand{\arraystretch}{1.10}
\begin{adjustbox}{width=\columnwidth}
\begin{tabular}{l|cccccc}
\toprule

\textbf{Report} 
& \textbf{Rev.} 
& \textbf{EBITDA}
& \textbf{Op. Inc.}
& \textbf{Net Inc.}
& \textbf{FCF} 
& \textbf{EPS} \\
\midrule

\multicolumn{7}{c}{LLMs + Web Search} \\

\hline 
gpt4o-mini 
& \uline{18.61}
& 74.80
& 81.89
& 44.31
& 67.10
& 51.36 \\

gpt4o 
& 20.36
& 80.72
& 85.70
& 66.12
& \uline{12.62}
& 51.88 \\

sonar-pro
& 22.12
& 78.92
& 49.91
& 42.98
& 27.95
& \textbf{20.79} \\

sonar-reason-pro
& 27.39
& 43.47
& 49.54
& 53.03
& 67.36
& 40.04 \\

gemini-2.5-flash
& \textbf{16.75}
& 23.78
& 29.70
& 30.35
& 29.12
& \uline{24.31} \\

\hline
\multicolumn{7}{c}{Deep Research Agents} \\
\hline

OpenAI
& 24.73
& \textbf{12.13}
& \textbf{16.78}
& \textbf{16.62}
& 18.54
& 39.06 \\

Gemini 
& 22.63
& \uline{22.52}
& \uline{29.18}
& 21.59
& \textbf{9.51}
& 27.23 \\

Perplexity
& 23.92
& 38.60
& 29.90
& \uline{18.76}
& 15.64
& 33.60 \\

\bottomrule
\end{tabular}
\end{adjustbox}

\caption{
Financial forecast accuracy measured by SMAPE (\%). Lower values indicate better performance.
Bold indicates best result per metric; underlining indicates second-best. Reports were generated using a prompt
template based on Firm A dataset.}
\label{tab:llm_web_fin_forecast_smape_a}
\end{table}

\begin{table}[htbp]
\centering
\small
\setlength{\tabcolsep}{6pt}
\renewcommand{\arraystretch}{0.90}
\begin{adjustbox}{width=0.65\columnwidth}
\begin{tabular}{l|cc}
\toprule
\textbf{Report}
& \textbf{Revenue} 
& \textbf{EPS} \\
\midrule
\multicolumn{3}{c}{LLMs + Web Search} \\
\hline

gpt4o-mini
& \textbf{16.56}
& 35.94 \\

gpt4o
& 17.40
& 40.51 \\

sonar-pro 
& 23.68
& 19.41 \\

sonar-reason-pro
& 18.12
& \uline{12.62} \\

gemini-2.5-flash
& 22.62
& 21.91 \\

\hline
\multicolumn{3}{c}{Deep Research Agents} \\
\hline

OpenAI
& 20.35
& 34.61 \\

Gemini 
& \uline{16.87}
& 23.75 \\

Perplexity
& 23.64
& \textbf{10.28} \\

\bottomrule
\end{tabular}
\end{adjustbox}

\caption{
Financial forecast accuracy measured by SMAPE (\%).
Lower values indicate better performance. Bold indicates best result per metric; underlining indicates second-best. Reports were generated using a prompt template based on Firm B dataset.
}
\label{tab:llm_web_fin_forecast_smape_b}
\end{table}

\noindent\textbf{Quantitative Evaluation - Stock Valuation.}  
Tables \ref{tab:llm_web_stock_price_3m6m_a} and \ref{tab:llm_web_stock_price_3m6m_b} summarize 3--6 month target price accuracy under Firm A and Firm B prompts.

On \textbf{Firm A}, deep research (DR) agents achieved the strongest hit rates (e.g., OpenAI 33.3\% at 3m; Gemini/Perplexity $\sim$57\% at 6m). However, \texttt{gpt4o} with web search delivered the lowest 6m MAE (12.5\%) and near-best bias (+4.9\%), indicating superior magnitude calibration at longer horizons. Overall, DR agents led on directional success, while web-enabled models were more precise on 6m error.

On \textbf{Firm B}, DR agents showed more stable cross-metric performance. OpenAI combined solid hit rates (31.8\% at 3m; 52.4\% at 6m) with low MAE and the lowest 6m bias (+4.1\%). Although \texttt{sonar-reason pro} achieved perfect hit rates, most web-search models exhibited larger optimism bias and weaker 6m consistency. Overall, DR agents were better calibrated on Firm B.

\begin{table}[htbp]
\centering
\scriptsize
\setlength{\tabcolsep}{4pt}
\renewcommand{\arraystretch}{1.1}
\begin{adjustbox}{width=\columnwidth}
\begin{tabular}{l|c|ccc}
\toprule
\multirow{2}{*}{\textbf{Report}} & \multirow{2}{*}{\textbf{Horizon}} & \textbf{Hit Rate}  & \textbf{MAE} & \textbf{Mean Bias} \\
                &                  &    (\%)            &   (\%)       &   (\%) \\
\hline
\multicolumn{5}{c}{LLMs + Web Search} \\
\hline

\multirow{2}{*}{gpt4o-mini}
& 3m & 17.6 & 24.1 & +18.8 \\
& 6m & 17.6 & \uline{17.4} & +11.0 \\
\hline

\multirow{2}{*}{gpt4o}
& 3m & 26.3 & \textbf{17.4} & +12.0 \\
& 6m & 47.4 & \textbf{12.5} & \uline{+4.9} \\
\hline

\multirow{2}{*}{sonar-pro}
& 3m & 15.4 & 51.3 & +35.2 \\
& 6m & 23.1 & 51.9 & +23.3 \\
\hline

\multirow{2}{*}{sonar-reason-pro}
& 3m & 0.0 & 90.4 & \textbf{+4.4} \\
& 6m & 10.0 & 91.0 & -10.5 \\
\hline

\multirow{2}{*}{gemini-2.5-flash}
& 3m & 17.6 & 27.1 & +15.3 \\
& 6m & 11.8 & 27.1 & +8.9 \\
\hline

\multicolumn{5}{c}{Deep Research Agents} \\
\hline

\multirow{2}{*}{OpenAI} 
& 3m & \textbf{33.3} & 26.4 & \uline{+10.0} \\
& 6m & 46.2 & 27.4 & \textbf{+4.3} \\ \hline

\multirow{2}{*}{Gemini}
& 3m & \uline{31.8} & 28.2 & +17.0 \\
& 6m & \textbf{57.1} & 28.0 & +8.4 \\ \hline

\multirow{2}{*}{Perplexity}
& 3m & 29.3 & \uline{22.3} & +19.7 \\
& 6m & \uline{56.4} & 19.1 & +11.0 \\

\bottomrule
\end{tabular}
\end{adjustbox}
\caption{
Target price metrics by horizon (3m and 6m only) using Firm A prompt style.
Hit rate: higher is better.
MAE: lower is better.
Mean bias ranking uses absolute magnitude (closer to zero is better) while retaining sign.
Bold indicates best and underlined indicates second-best within each horizon and metric.
}
\label{tab:llm_web_stock_price_3m6m_a}
\end{table}

\begin{table}[htbp]
\centering
\scriptsize
\setlength{\tabcolsep}{4pt}
\renewcommand{\arraystretch}{1.1}
\begin{adjustbox}{width=\columnwidth}
\begin{tabular}{l|c|ccc}
\toprule
\multirow{2}{*}{\textbf{Report}} & \multirow{2}{*}{\textbf{Horizon}} & \textbf{Hit Rate} & \textbf{MAE} & \textbf{Mean Bias} \\
                &                  & (\%)              & (\%)         &   (\%) \\
\hline
\multicolumn{5}{c}{LLMs + Web Search} \\
\hline

\multirow{2}{*}{gpt4o-mini}
& 3m & 6.2 & 38.7 & +35.3 \\
& 6m & 0.0 & 30.3 & +29.0 \\
\hline

\multirow{2}{*}{gpt4o}
& 3m & 0.0 & 29.5 & +28.8 \\
& 6m & 0.0 & 27.1 & +24.3 \\
\hline

\multirow{2}{*}{sonar-pro}
& 3m & \uline{33.3} & 18.1 & \uline{+14.9} \\
& 6m & 0.0 & 19.8 & +19.8 \\
\hline

\multirow{2}{1.5cm}{sonar-reason pro}
& 3m & \textbf{100.0} & \uline{17.8} & +17.8 \\
& 6m & \textbf{100.0} & \textbf{14.4} & +12.4 \\
\hline

\multirow{2}{*}{gemini-2.5-flash}
& 3m & 15.0 & 61.6 & +42.1 \\
& 6m & 10.0 & 59.1 & +37.2 \\
\hline

\multicolumn{5}{c}{Deep Research Agents} \\
\hline

\multirow{2}{*}{OpenAI}
& 3m & 31.8 & \textbf{16.3} & \textbf{+10.7} \\
& 6m & \uline{52.4} & 17.3 & \textbf{+4.1} \\ \hline

\multirow{2}{*}{Gemini}
& 3m & 22.3  &19.8 &+19.8 \\
& 6m & 47.6 & \uline{16.0} &+13.1 \\ \hline

\multirow{2}{*}{Perplexity}
& 3m & 20.9 & 18.1 & +16.0 \\
& 6m & 43.5 & 17.5 & \uline{+9.3} \\

\bottomrule
\end{tabular}
\end{adjustbox}
\caption{
Target price metrics by horizon (3m and 6m only) using Firm B prompt style.
Hit rate: higher is better.
MAE: lower is better.
Mean bias ranking uses absolute magnitude (closer to zero is better) while retaining sign.
Bold indicates best and underlined indicates second-best within each horizon and metric.
}
\label{tab:llm_web_stock_price_3m6m_b}
\end{table}

\noindent\textbf{Quantitative Evaluation - Investment Recommendation.}  
Tables \ref{tab:llm_web_stock_rec_a} and \ref{tab:llm_web_stock_rec_b} report directional accuracy and SRL for 3--6 month recommendations.

On \textbf{Firm A}, LLMs with web search outperformed deep research (DR) agents on directional accuracy. \texttt{gpt4o} achieved the highest accuracy (84.2\% at both horizons) with low SRL, while \texttt{gpt4o-mini} was close behind. Although DR agents were competitive on SRL (e.g., Gemini 0.031 at 3m; OpenAI 0.052 at 6m), they generally trailed in directional accuracy.

On \textbf{Firm B}, the pattern reversed. DR agents, particularly OpenAI, led in directional accuracy (60.9\% at 3m; 78.3\% at 6m). Web-search models achieved lower SRL at 6m (e.g., \texttt{gpt4o-mini} 0.047), but their directional performance was weaker overall. In summary, web-enabled models were stronger under Firm A prompts, whereas DR agents held an advantage under Firm B.

\begin{table}[htbp]
\centering
\scriptsize
\setlength{\tabcolsep}{4pt}
\renewcommand{\arraystretch}{1.1}
\begin{adjustbox}{width=0.9\columnwidth}
\begin{tabular}{l|c|cc}
\toprule
\textbf{Report} & \textbf{Horizon} & \textbf{Dir. Acc. (\%)} & \textbf{SRL} \\

\hline
\multicolumn{4}{c}{LLMs + Web Search} \\
\hline

\multirow{2}{*}{gpt4o-mini}
& 3m & \uline{82.4} & 0.053 \\
& 6m & \uline{82.4} & 0.066 \\
\hline

\multirow{2}{*}{gpt4o}
& 3m & \textbf{84.2} & 0.046 \\
& 6m & \textbf{84.2} & \uline{0.056} \\
\hline

\multirow{2}{*}{sonar-pro}
& 3m & 53.8 & 0.228 \\
& 6m & 69.2 & 0.236 \\
\hline

\multirow{2}{1.5cm}{sonar-reason pro}
& 3m & 10.0 & 0.912 \\
& 6m & 10.0 & 0.901 \\
\hline

\multirow{2}{*}{gemini-2.5-flash}
& 3m & 70.6 & \uline{0.038} \\
& 6m & 76.5 & \uline{0.056} \\
\hline

\multicolumn{4}{c}{Deep Research Agents} \\
\hline

\multirow{2}{*}{OpenAI} 
& 3m & 80.0 & 0.046 \\
& 6m & 66.7 & \textbf{0.052} \\ \hline

\multirow{2}{*}{Gemini}
& 3m & 70.6 & \textbf{0.031} \\
& 6m & 76.5 & 0.058 \\ \hline

\multirow{2}{*}{Perplexity} 
& 3m & 69.6  & 0.092 \\
& 6m & 72.7  & 0.105 \\ 

\bottomrule
\end{tabular}
\end{adjustbox}

\caption{
Stock recommendation metrics by horizon (3m and 6m only) based on Firm A prompt style.
Directional accuracy: higher is better.
SRL: lower is better.
Bold indicates best and underlined indicates second-best within each horizon and metric.
}
\label{tab:llm_web_stock_rec_a}
\end{table}

\begin{table}[htbp]
\centering
\scriptsize
\setlength{\tabcolsep}{4pt}
\renewcommand{\arraystretch}{1.1}
\begin{adjustbox}{width=0.9\columnwidth}
\begin{tabular}{l|c|cc}
\toprule
\textbf{Report} & \textbf{Horizon} & \textbf{Dir. Acc. (\%)} & \textbf{SRL} \\

\hline
\multicolumn{4}{c}{LLMs + Search} \\
\hline

\multirow{2}{*}{gpt4o-mini}
& 3m & 52.4 & \textbf{0.033} \\
& 6m & \uline{76.2} & \textbf{0.047} \\
\hline

\multirow{2}{*}{gpt4o}
& 3m & \uline{55.0} & \uline{0.035} \\
& 6m & 70.0 & \uline{0.054} \\
\hline

\multirow{2}{*}{sonar-pro}
& 3m & 18.8 & 0.463 \\
& 6m & 18.8 & 0.483 \\
\hline

\multirow{2}{1.5cm}{sonar-reason pro}
& 3m & 20.0 & 0.619 \\
& 6m & 25.0 & 0.614 \\
\hline

\multirow{2}{*}{gemini-2.5-flash}
& 3m & 40.0 & 0.329 \\
& 6m & 45.0 & 0.344 \\
\hline

\multicolumn{4}{c}{Deep Research Agents} \\
\hline 

\multirow{2}{*}{OpenAI} 
& 3m & \textbf{60.9} & 0.061 \\
& 6m & \textbf{78.3} & 0.090\\ \hline

\multirow{2}{*}{Gemini} 
& 3m & 41.7 & 0.218\\
& 6m & 54.2 & 0.245 \\ \hline

\multirow{2}{*}{Perplexity}
& 3m & 40.9 & 0.181 \\
& 6m & 63.6 & 0.192 \\

\bottomrule
\end{tabular}
\end{adjustbox}

\caption{
Stock recommendation metrics by horizon (3m and 6m only) using Firm B prompt style.
Directional accuracy: higher is better.
SRL: lower is better.
Bold indicates best and underlined indicates second-best within each horizon and metric.
}
\label{tab:llm_web_stock_rec_b}
\end{table}











\subsection{Verifiability Analysis}
\label{sec:app_}

\begin{table}[htbp]
\centering
\small
\setlength{\tabcolsep}{6pt}
\renewcommand{\arraystretch}{1.05}
\begin{adjustbox}{width=\columnwidth}
\begin{tabular}{
    l
    S[table-format=2.2]
    S[table-format=2.2]
    S[table-format=2.2]
    S[table-format=2.2]
}
\toprule
\multicolumn{1}{l}{Primary Category} &
\multicolumn{1}{c}{OpenAI} &
\multicolumn{1}{c}{Gemini} &
\multicolumn{1}{c}{Grok} &
\multicolumn{1}{c}{Perp} \\
\midrule
Market Price Trading               & 0.00  & 7.41 & 1.61 & 0.00 \\
Market Cap Valuation               & 12.50 & 8.64 & 2.42 & 4.35 \\
Financial Statements               & 15.62 & 13.58 & 23.39 & 23.91 \\
Profitability Ratios               & 12.50 & 6.17 & 10.48 & 7.61 \\
Growth Change Rates                & 3.12  & 7.41 & 17.74 & 9.78 \\
Capital Return Actions             & 9.38  & 8.64 & 8.87 & 14.13 \\
Operational Segment Mix            & 9.38  & 6.17 & 20.16 & 18.48 \\
Guidance Forecast Targets          & 3.12  & 17.28 & 13.71 & 15.22 \\
ESG Ratings Scores                 & 31.25 & 18.52 & 0.81 & 5.43 \\
Macro Industry Metrics             & 0.00  & 0.00 & 0.00 & 0.00 \\
Date Time Point                    & 3.12  & 6.17 & 0.81 & 1.09 \\
\bottomrule
\end{tabular}
\end{adjustbox}
\caption{Provider-wise hallucination distribution (\%) across numerical categories}
\label{tab:provider_primary_category_percent_transposed}
\end{table}

We further analyze hallucinations on quantitative claims by dividing the quantitative hallucinations into the following subtypes:  Market Price Trading, Market Cap Valuation, Financial Statements, Profitability Ratios, Growth Change Rates, Capital Return Actions, Operational Segment Mix, Guidance Forecast Targets, ESG Ratings Scores, Macro Industry Metrics, and Date Time Point. We briefly describe each subtype below: 

\noindent\textbf{Market Price Trading}: Anything related to a security’s trading behavior or price action (price moves, volume, volatility, etc.).

\noindent\textbf{Market Cap Valuation}: Company value and valuation metrics (market cap, enterprise value, P/E, EV/EBITDA, etc.).

\noindent\textbf{Financial Statements}: Statement-line numbers and accounting items (revenue, COGS, operating income, net income, EPS, etc.), usually reported from 10-K/10-Q/earnings.

\noindent\textbf{Profitability Ratios}: Margin/return ratios and profitability rates derived from statements (gross margin, operating margin, net margin, etc.).

\noindent\textbf{Growth Change Rates}: Any “change over time” metric (YoY/QoQ growth, CAGR, percent increase/decrease, etc.).

\noindent\textbf{Capital Return Actions}: Shareholder return actions and structural changes (dividends, buybacks, splits, etc.).

\noindent\textbf{Operational Segment Mix}: Business composition and operating metrics by segment/product/region/customer (segment revenue, mix shift, etc.).

\noindent\textbf{Guidance Forecast Targets}: Forward-looking numbers or targets (company guidance, analyst forecasts, price targets,etc.).

\noindent\textbf{ESG Ratings Scores}: ESG-related ratings and scored assessments (MSCI ESG rating, Sustainalytics risk score, etc.) and similar quantified ESG evaluations.

\noindent\textbf{Macro Industry Metrics}: Broader economy/industry indicators (inflation, rates, unemployment, GDP, etc.).

\noindent\textbf{Date Time Point}: Pure timing references or point-in-time qualifiers used to anchor claims (as-of dates, quarter/year references, etc.).

\subsection{Professionals vs DR Agents Analyses}
\label{sec:prof_vs_dr_analyses}

\subsubsection{Qualitative Research Quality}
\label{sec:quali_analysis_discuss}

\textbf{Comprehensiveness.} Our analysis found deep research agents generated research reports containing nearly 100\% of sections instructed by the prompt. Superficially, DR reports seem to cover the same key points and discussions as professional reports.  In a contrast of professional and DR reports of Apple Inc. below, the professional report is able to cover the current development of key products like iPhone 17, their sales and granular revenue projections. Their research consider the uniqueness in each company and sector. Meanwhile, all the DR reports show a common pattern of applies a generic template of forecasting financials across all companies omitting company-specific and industry-specific KPIs.  Moreover, DR reports also failed to make assumptions to different scenarios (see the assumption discussion for more details) which is a key method to navigate volatile markets and ensure that investment recommendations are robust. 

\begin{quoteblock}{QuoteGreen}
    \small
    \textsc{Professional Report}
    
    \vspace{5pt}
    
    ``We expect digestion of pull forward demand as well as expected price increases with \textbf{iPhone 17} lineup to drive the company's guidance of flat revenue growth ... 

    \begin{center}
    \begin{tabular}{ l|ccc}
    Year & 2023A & 2024A & 2025E \\ 
    \hline
    iPhone Units (M) & 224 & 229 & 230 \\  
    iPhone Revenue (\$B) & 200.6 & 201.2  & 195.9 \\
    ... & & & \\
    Revenue (\$B) & 383.3 & 391.0 & 400.1 \\
    \end{tabular}
    \end{center}
    ''
\end{quoteblock}

\begin{quoteblock}{QuoteRed}
    \small

    \textsc{DR Report}
    
    \vspace{5pt}
    ``
    Revenue growth modeling assumes \textbf{iPhone} sales stabilization around current levels with modest annual declines ...
    
    \vspace{5pt}

    \begin{center}
    \begin{tabular}{ l|ccc}
    Year & 2023A & 2024A & 2025E \\ 
    \hline
    Revenue (\$B)   & 383.3 & 391.0 & 405.2 \\
    Op. Margin (\%) & 29.8  & 31.5  & 30.8  \\
    \end{tabular}
    \end{center}    
    ''

\end{quoteblock}

\noindent\textbf{Coherence}. Although the coherence within the same section of a report is acceptable, DR reports show notable inconsistency across sections. For example, AI made an assumption of 8\% revenue CAGR in a report below. From the detailed table of revenue projection in another section, however, the calculation indicates this metric should $(250/128.7)^{1/7} =9.9\%$, not 8\%. 
\begin{quoteblock}{QuoteRed}
    \small

    \textsc{DR Report}
    
    \vspace{5pt}
    
    ``Key inputs: \textbf{Revenue CAGR ~8\% (2022–29)}, driven by loan and fee growth as rates stay mildly elevated ...
    
    \vspace{5pt}
    
    | Year | Revenue (US\$B) |\\
    |2022A | \textbf{128.7} | \\ 
    ... \\
    |2029E | \textbf{250.0} | \\    
    ''
\end{quoteblock}

In another example below, a DR report used 32-34\% operating margins projection between 2025 and 2029 for stock valuation. But its financial model forecasted 35.1\% and 35.3\% in 2028 and 2029. This is a obvious contradiction.  

\begin{quoteblock}{QuoteRed}
    \small

    \textsc{DR Report}
    
    \vspace{5pt}
    
    ``The valuation reflects revenue compound annual growth rates of 8-10\% \textbf{over the next five years}. Operating margins are projected to \textbf{stabilize 32-34\%}...
    
    \vspace{5pt}
    | Year | Op Margin \% |\\
    |2025E | 33.5 | \\ 
    ... \\
    |2028E | \textbf{35.1} | \\
    |2029E | \textbf{35.3} | \\    
    ''
\end{quoteblock}
We can barely detect coherence issue in professional reports thought. A reasonable guess is DR agents usually perform research of each section in parallel and the alignment of same metric/argument is not strictly enforced among sections.

\noindent\textbf{Assumption}. Solid analyses of financial projection and stock valuation rely on well validated assumptions of key financial metrics like revenue growth and gross margin etc. In an example of professional reports below, analysts well reasoned their assumption of 7\% compound revenue growth of Apple Inc from 2025 to 2029. Their logic is that iPhone sale as the main revenue source will grow steadily 6\% with modest unit price increase. In some of DR reports (see an example below), however, such assumptions were made with relatively weak reasoning. No further logical discussion of why "service growth will be high single-digit".  

\begin{quoteblock}{QuoteGreen}
    \small
    \textsc{Professional Report}
    
    \vspace{5pt}
    
    ``We project \textbf{7\% compound annual revenue growth} through fiscal 2029. The \textbf{iPhone} will be the \textbf{most significant contributor} to revenue over our forecast, and we project \textbf{6\% growth for iPhone revenue} over the next five years. We expect this to be driven primarily by \textbf{unit sales growth}, with \textbf{modest pricing increase}.
    ''
\end{quoteblock}

\begin{quoteblock}{QuoteRed}
    \small

    \textsc{AI Report}    

    \vspace{5pt}
    
    ``
    Key Valuation Assumptions:

    \vspace{5pt}
    
    \begin{tabular}{ lll}
    Assumption & Value & Comment \\ \hline
    Revenue      &       &  \textbf{Services growth high}   \\
     CAGR        & ~4\%  &  \textbf{single-digit}; hardware \\ 
    (FY24–FY29)  &       &  low growth.
    \end{tabular}

    ''

       
\end{quoteblock}

Prudential financial research should evaluate multiple potential futures and help investors navigate the volatile financial markets and economy conditions. Professional analysts routinely provide quantified scenario analysis with bull/bear cases. In an example of professional report below, analysts argued the fairly valued stock price can bring down \$175 due to antitrust impact to revenue; on the other hand, the price may reach \$266 at bull case since the dominance of online search market.  while AI reports consistently deliver single-point estimates (\$266 in 12 months) without meaningful sensitivity ranges.

\begin{quoteblock}{QuoteGreen}
    \small
    \textsc{Professional Report}
    
    \vspace{5pt}
    
    ``The bear case includes a material deterioration in the firm’s search revenue stemming from antitrust regulatory action... Our fair value estimate in this \textbf{\color{black}hypothetical bear case is \$175 per share}.
    
    \vspace{5pt}
    
    The bull case for [company x] envisions a scenario where its search maintains its dominance in the online search market ... The fair value estimate in the \textbf{\color{black}bull-case scenario is \$266 per share}.''
\end{quoteblock}

\begin{quoteblock}{QuoteRed}
    \small

    \textsc{DR Report}
    
    \vspace{5pt}
    
    ``Synthesizing our DCF intrinsic value of \$213 and the potential for some multiple expansion as the company executes on its AI strategy ..., we arrive at a \textbf{\color{black}12-month price target of \$210}.''
\end{quoteblock}

\noindent\textbf{Analytical Depth.}
The analytical discussions in equity research reports reflect analysts' independent opinions on what a stock is performing and what is a right investment strategy. Such in-depth analysis was clearly demonstrated in a professional reports below. Analysts presented the logic of data center (DC) revenue of a company is up 73\%. This massive DC growth will lead a significant gross margin expansion. Given the assumption of gross margin boost, the current stock price is fairly valued. Most of DR reports as we examined only have relatively simple analyses.  They are more descriptive rather than analytical, missing sound causal analysis. In addition, most of these analyses synthesize the logic from others' research in the references. Hence, consistency is a concern. They cannot be treated as original opinions of professional analysts where investor seek insights from.  

\begin{quoteblock}{QuoteGreen}
    \small
    \textsc{Professional Report}
    
    \vspace{5pt}
    
    ``\textbf{Data center (DC) revenue} of \$39.1 billion was \textbf{up 73\%} year over year ... 
    \vspace{5pt}
    
    \textbf{Massive DC growth} has been margin-accretive, as \textbf{gross margin expanded} from 57\% in 2023 to 75\% in 2025. In the long run, we model a 74\% \textbf{gross margin in 2027}...

    \vspace{5pt}
    \textbf{Stock price} now appears fairly valued to us assume ... boost long-term \textbf{gross margin assumption}...
    ''
    
\end{quoteblock}

\begin{quoteblock}{QuoteRed}
    \small

    \textsc{DR Report}
    
    \vspace{5pt}
    
    ``Revenue of \$44.1 billion exceeded expectations, \textbf{driven primarily by Data Center} segment performance of \$39.1 billion.. 
    
    \vspace{5pt}
    \textbf{Target margins of 72-75\% }incorporate mix shifts toward \textbf{higher-value data center products}...
    ''
\end{quoteblock}

\subsubsection{Quantitative Research Quality}
\textbf{Financial Projection}. There are notable differences on financial modeling approaches between professionals and deep research reports. Human analysts build forecasts from the product/service segments and quarters (see a contrast example in section \ref{sec:quali_analysis_discuss}) level up. Meanwhile, all DR reports only generate forecasts of company wide annual totals. Analysts' approach offers superior granularity, accuracy, and actionable insights. In addition, human analysts systematically discussed the management guidance figures from earnings calls and then adjust their own estimates based on guidance credibility. Analysts also tie their forecasts to key revenue impact events (see the modeling of revenue impacts by "potential new PC CPU business" and "lifting H20 ban to China"). These practices, however, are missed in DR generated reports. As a results, we can observe from Table \ref{tab:forecast_accuracy_smape_all} professionals can produce more accurate predictions of financial items than DR agents. 

\begin{quoteblock}{QuoteGreen}
    \small
    \textsc{Professional Report}
    
    \vspace{5pt}
    
    ``... may introduce a \textbf{PC CPU} in fiscal 2027, \textbf{boosting revenue} in this segment (if such revenue is in fact reported here) to \textbf{\$22 billion}. We model \textbf{10\%} average annual revenue growth in gaming thereafter ...
    ''
    \vspace{10pt}
    
    ``... \textbf{\$10 billion} higher than our prior assumptions thanks to the US reversal of its ban on \textbf{H20 sales} into \textbf{China}. With China coming back into the model, perhaps bigger than ever, we model \textbf{35\% growth}...''
    
\end{quoteblock}

\noindent\textbf{Stock Valuation and Recommendation}. The accuracy of financial projections impact the correctness of the stock valuation because the price model is built upon the company's financial performance. All the professional reports used discounted cash flow model (DCF), which is an industry standard approach, to estimates a company’s intrinsic value \cite{olbert2025industry}. Professionals give a step-by-step analysis of their discounted cash flow (DCF) models fed by detailed industry and company assumptions, augmented by scenario analysis and competitive assessment. While DR reports provide limited analysis and especially lack a stage-by-stage present value breakdown. As illustrated in section \ref {sec:quali_analysis_discuss}, professional analysts have dedicated discussions on what are bull or bear scenarios due to the uncertainty and how these scenarios can impact to the stock valuation. Some analysts even quantify the discount or premium of threshold for buy or sell recommendation given different uncertainty ratings (see an example below). We were not able to observe the similar level of analyses in DR reports.

\begin{quoteblock}{QuoteGreen}
    \small

    \begin{tabular}{ lll}
    Uncertainty Level & Buy Threshold & Sell Threshold \\ \hline
    Low	              & 18\% discount   &20\% premium  \\
    Medium            & 28\% discount   &30\% premium  \\
    High              & 38\% discount   &40\% premium  \\
    \end{tabular}

\end{quoteblock}

\subsubsection{Verifiability and Credibility}

\noindent\textbf{Verifiability.} Although many deep research agents implemented internal algorithms to ensure the factuality of generated content, there are still some degree (10-20\%) of claims and facts that cannot be verified based on their citations. The most common reason is hallucinations exist stickily. In an example below, the paragraph claims "investigation was closed in February 2024" and cites a UK government page about this case. However, the content of this page actually indicates this investigation was closed on November 3. Clearly, this was a hallucination of key dates.       

\begin{quoteblock}{QuoteRed}
    \small

    \textsc{DR Report}
    
    \vspace{5pt}
    Claim: ``In \textbf{February 2024}, the UK CMA accepted commitments from Amazon regarding marketplace practices and \textbf{closed its investigation} [1].”
    
    \vspace{8pt}
    
    [1] https://www.gov.uk/cma-cases/investigation-into-amazons-marketplace
    
    \vspace{10pt}
    
    Evidence from Citation: 
    
    ...\\
    \vspace{5pt}
    Case Timetable:
    
    \vspace{5pt}
    
    \begin{tabular}{l|l}
    Date & Action \\ \hline
    \textbf{3 November} & CMA commitments decision  \\
               & published and investigation closed
    \end{tabular}
    
\end{quoteblock}

In other example, the claim of "Synopsys's acquisition of Ansys at \$35B" itself is true based on other sources. However, its citing source cannot be found. We double checked its link on wayback machine \footnote{https://web.archive.org/}, which a digital archive of world wide web, and can confirm it never existed before. It is an obvious indication of hallucination on the referencing sources.

\begin{quoteblock}{QuoteRed}
    \small
    \textsc{DR Report}
    
    \vspace{5pt}
    Claim: ``Synopsys’ acquisition of Ansys is valued at approximately \$35 billion [1]” 

    \vspace{8pt}
    
    [1] https://www.prnewswire.com/synopsys-ansys-acquisition

    \vspace{10pt}
    Evidence from citation: 

    ...\\
    404 not found
    
\end{quoteblock}

In another example, the claim of "Amazon had approximately 10.6 billion shares outstanding" is supported by additional evidence, but not supported by provided citation.

\begin{quoteblock}{QuoteRed}
    \small
    \textsc{DR Report}
    
    \vspace{5pt}
    Claim: ``Amazon had approximately 10.6 billion shares outstanding [1]” 

    \vspace{8pt}
    
    [1] https://futurumgroup.com/insights/amazon-q1-fy-2025-earnings

    \vspace{10pt}
    Evidence from citation: 

    ...\\
    No mention of outstanding shares in the citations.
    
\end{quoteblock}

Overall, the aforementioned issues of verifiability and credibility of the claims and their sources may bring serious doubts from professionals to use AI generated reports in their research tasks. 


\noindent\textbf{Credibility}. In this section, we analyze two different citation examples, one from a high-trust source and the other from a low-trust source. First, for the claim ``In fiscal Q1 2025, Apple's Services net sales were \$26.34 billion.", we got a citation from sec.gov, which is a high-trust source.
\begin{quoteblock}{QuoteRed}
    \small
    \textsc{DR Report}
    
    \vspace{5pt}
    Claim: In fiscal Q1 2025, Apple's Services net sales were \$26.34 billion.

    \vspace{8pt}
    
    Web-link: https://www.sec.gov/Archives/edgar/data/
    320193/000032019325000008/aapl-20241228.
    htm\#:~:text=Net\%20sales\%3A\%20\%20,119\%2C575

\end{quoteblock}

While for the following claim ``Amazon’s advertising revenue grew 18\% in Q1 2025.", the web-source is based on social-media x, which is a low-trust source.

\begin{quoteblock}{QuoteRed}
    \small
    \textsc{DR Report}
    
    \vspace{5pt}
    Claim: Amazon’s advertising revenue grew 18\% in Q1 2025.

    \vspace{8pt}
    
    Web-link: https://x.com/fiscal\_ai/status/191803646482022
    0252

\end{quoteblock}

\subsubsection{Verifiability of Human Reports}
\label{sec:human_report}
In addition to verifying DR agent-generated reports, we also verify human-generated reports from firms A and B. Verifying human-generated reports is more challenging because they often lack citations. Hence, we extract the claims from the reports first using GPT-5.2-reasoning, then use the same model to verify each claim. For claim verification, we prompt GPT-5.2-reasoning to search for web resources to verify each claim. For this experiment, we manually feed the reports to ChatGPT for verification. For Firm A, we find $F(R)$ of 75.93\%, $NV(R)$ of 19.35\%, and error rate $E(R)$ of 4.72\%. We further investigated some of the claims classified as error and found that they are generally classified as error either due to rounding or due to conflicting evidence. Hence, claims classified as errors by an LLM might not be incorrect.
Additionally, for Firm B, $F(R)$ is 51.08\%, $NV(R)$ is 2.32\% and $E(R)$ is 2.32\%.

In the following two examples, human reports mentioned two numerical claims, where LLM found evidence that the difference between reported values and evidence values exceeds the 1\% tolerance limit.

\begin{quoteblock}{QuoteRed}
    \small
    \textsc{Human Report}
    
    \vspace{5pt}
    Claim: Johnson \& Johnson acquired Momenta Pharmaceuticals for \$6 billion.

    \vspace{8pt}
    
    Web-link: https://www.sec.gov/Archives/edgar/data/
    1235010/000110465920096687/tm2028912d1\_ex99-1.htm

    \vspace{10pt}
    Error Reason: 

    ...\\
    In-date filing/press release materials state an implied equity value of \$6.5B, not \$6.0B (difference >1\%). Post-dated sources ignored.
    
\end{quoteblock}

\begin{quoteblock}{QuoteRed}
    \small
    \textsc{Human Report}
    
    \vspace{5pt}
    Claim: The report's financial table shows CVS operating income of 14,599 USD million for fiscal year 2023. 

    \vspace{8pt}
    
    Web-link: https://cvshealth2023inreview.com/financial-highlights/

    \vspace{10pt}
    Error Reason: 

    ...\\
    CVS reports FY2023 operating income of 13,743 USD in this release (not 14,599 USD); the difference exceeds the 1\% tolerance.
    
\end{quoteblock}


\end{document}